\PassOptionsToPackage{pdftex}{graphicx}
\documentclass[preprint,review,12pt]{elsarticle}

\usepackage[utf8]{inputenc}
\usepackage{booktabs} 
\usepackage{tabularx} 
\usepackage{threeparttable} 
\usepackage{multirow}
\usepackage{xspace}
\usepackage{url}
\usepackage{amssymb}
\usepackage{scalefnt}
\usepackage{footnote}
\usepackage{array}
\usepackage[table]{xcolor}
\makesavenoteenv{table}
\makesavenoteenv{tabular}
\usepackage[framemethod=tikz]{mdframed}
\usepackage{booktabs}
\usepackage[utf8]{inputenc}
\usepackage{subcaption}
\usepackage{times}
\usepackage{graphicx}
\usepackage{amsmath}
\usepackage{rotating}
\usepackage{setspace}
\usepackage{hhline}
\usepackage{rotating}

\usepackage{lineno}

\usepackage{color} 
\usepackage{soul} 

\usepackage{filecontents}
\setcitestyle{square}
\biboptions{sort&compress}

\usepackage{etoolbox}

\journal{Information Processing in Agriculture}

\begin{document}
\begin{frontmatter}




\title{Spectroscopy Analysis with Machine Learning Regression for the Quantification of Carbon and Nitrogen Contents in Inceptisol and Oxisol Soil Types: Comparing Different Preprocessing and Validation methods as well as Feature Importance }





\author[1]{Vinicius Herique Kieling}

\author[1]{Guilherme Macedo Baggio}

\author[2]{Felipe Augusto Bueno Rossi}

\author[3]{Marco Antonio de Castro Barbosa}

\author[1]{Dalcimar Casanova}

\author[2]{Larissa Macedo dos Santos Tonial}

\author[1]{Jefferson Tales Oliva}


\affiliation[1]{organization={Graduate Program in Electrical and Computer Engineering, Federal University of Technology - Paraná},
            addressline={Via do Conhecimento, s/n - KM 01 - Fraron}, 
            city={Pato Branco},
            postcode={85503-390}, 
            state={Paraná},
            country={Brazil}}

\affiliation[2]{organization={Academic Department of Chemistry, Federal University of Technology-Paraná},
            addressline={Via do Conhecimento, s/n - KM 01 - Fraron}, 
            city={Pato Branco},
            postcode={85503-390}, 
            state={Paraná},
            country={Brazil}}

\affiliation[3]{organization={Graduate Program in Production and Systems Engineering, Federal University of Technology-Paraná},
addressline={Via do Conhecimento, s/n - KM 01 - Fraron}, city={Pato Branco},
            postcode={85503-390}, 
            state={Paraná},
            country={Brazil}}

\begin{abstract}
Near-Infrared (NIR) spectroscopy has emerged as a promising alternative to traditional soil analysis methods, offering advantages such as speed, low cost, and non-destructive testing. This work proposes a machine learning (ML) approach to calibrate predictive models for carbon (C) and nitrogen (N) content in Oxisols and Inceptisols, utilizing NIR spectral data acquired with a portable MyNIR device. Various preprocessing methods were evaluated, with the most effective being the Savitzky-Golay (SG) filter and a robust outlier removal method based on the Nonlinear Iterative Partial Least Squares (NIPALS) algorithm coupled with a Huber loss function. Multiple validation strategies were compared, including 10-fold cross-validation, leave-one-out, and holdout via the Kennard-Stone method, followed by standardization. Stacking ensemble learning models were employed, using Partial Least Squares (PLS), Support Vector Regression (SVR), and Ridge as base models, with linear regression as the meta-model. The models were evaluated using $R^2$, Root Mean Squared Error (RMSE), Mean Absolute Error (MAE), and Ratio of Performance Deviation (RPD) metrics. The results indicated superior performance for Oxisols ($R^2 = 0.91$ for C and $R^2 = 0.89$ for N), while Inceptisols yielded satisfactory but less expressive results ($R^2 = 0.79$ for C and $R^2 = 0.77$ for N). The performance gap between soil types suggests the influence of pedological characteristics. Furthermore, the models achieved an RPD $> 2.0$ with low overfitting, validating the potential of this approach for rapid C and N quantification. This study contributes to the optimization of sustainable agricultural practices, aligning with the demand for efficient and environmentally friendly analytical methods. The developed technique enables faster decision-making for producers and consultants based on organic matter content, fertility indicators, and nutrient availability.
\end{abstract}



\end{frontmatter}

\section{Introduction}

Soil is indispensable in agriculture by providing essential nutrients for crop development, particularly nitrogen (N) and carbon (C), which are important indicators of soil fertility, quality, health, and sustainability \cite{nie2017detection}.

Soil Organic Matter (SOM), primarily composed of the C and conventionally estimated to contain approximately 58\% C  according to the Van Bemmelen factor (1.724) \cite{pribyl2010} , is a key component in soil fertility \cite{vanfertilidade} and soil chemistry \cite{sparks2003environmental,segnini2008estudo}, while also contributing to atmospheric C sequestration  and soil chemistry. {In this context, the behavior and distribution of SOM can vary considerably among soil classes due to differences in mineralogy, texture, and degree of weathering. }

The soil classes evaluated in this study were Oxisols and Inceptisols, according to the USDA Soil Taxonomy, corresponding to Latossolos and Cambissolos in the Brazilian Soil Classification System. These two types of soil are predominant in Brazil, covering almost 40~\% of the Brazilian territory. Oxisols, the most widespread in Brazil, are highly weathered, with low natural fertility, predominantly composed of  Fe and Al oxides (hematite, gibbsite), and characterized by low Cation Exchange Capacity (CEC). Inceptisols are less widespread throughout the country's territory. These are shallow and poorly developed, with an incipient B horizon, variable texture, moderate natural fertility, and high susceptibility to erosion, commonly occurring in sloping areas \cite{sibcs2025}.

\begin{figure}[!ht]
    \centering
    \begin{subfigure}[b]{0.48\textwidth}
        \centering
        \includegraphics[width=\textwidth]{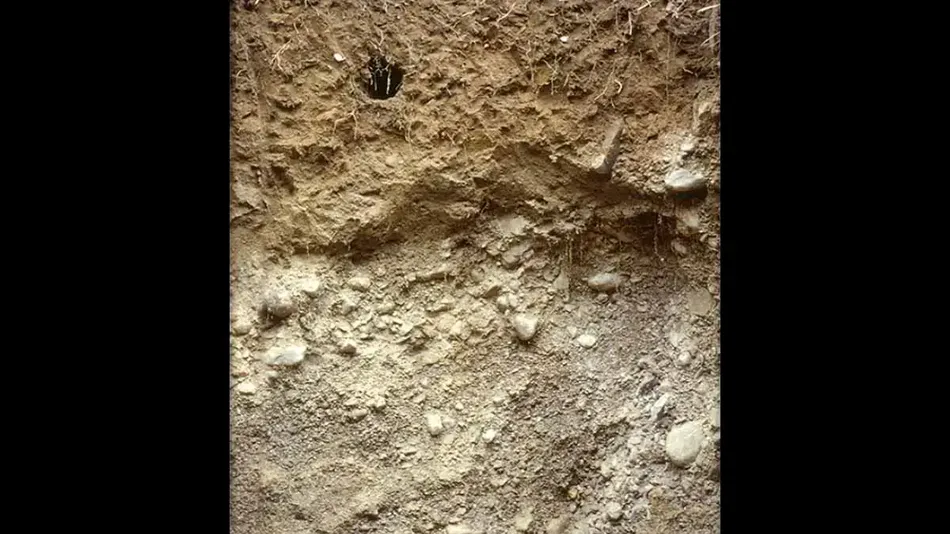}
        \caption{Inceptisol profile.}
        \label{fig:inceptisol}
    \end{subfigure}
    \hfill 
    \begin{subfigure}[b]{0.48\textwidth}
        \centering
        \includegraphics[width=\textwidth]{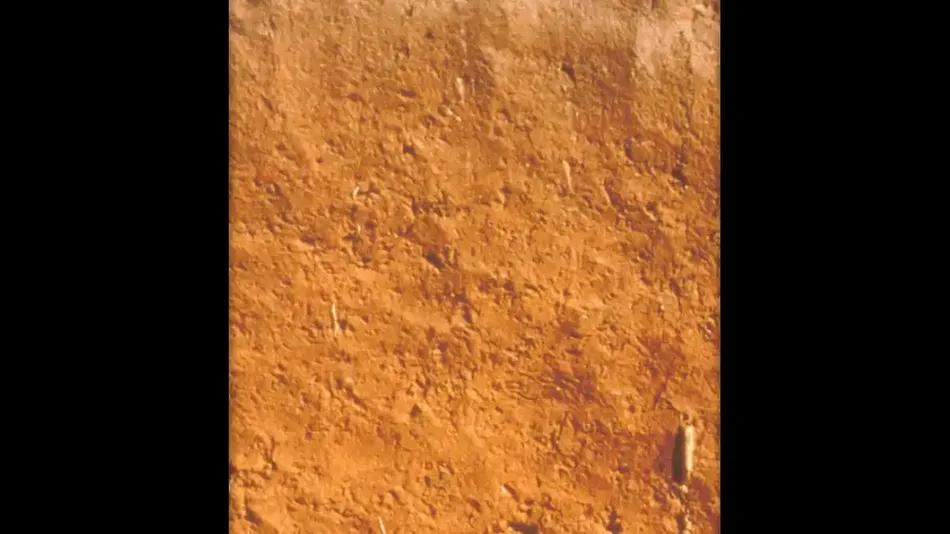} 
        \caption{Oxisol profile.}
        \label{fig:oxisol}
    \end{subfigure}
    
    \caption{Soil profiles analyzed in this study: (a) Inceptisol and (b) Oxisol. \cite{uidaho_inceptisols}}
    \label{fig:oxisolandinceptisol}
\end{figure}

Traditional soil analysis techniques, frequently employed by research and routine laboratories, present several limitations that can compromise the precision of the results \cite{rossel2016global}. As an example, the Walkley–Black method employed for soil organic C determination, may present incomplete oxidation of organic matter \cite{fernandes2015, shamrikova2022, nelson1996total}, necessitating the use of correction factors that may not accurately reflect reality. According to \citet{jones1991kjeldahl}, the method recovers only about 77\% of the C and employs a single correction factor (typically 1.33), which introduces error \cite{nelson1996total}. Furthermore, the method, interferences related to soil mineralogy, texture, and inorganic constituents, affecting the accuracy of C quantification \cite{fernandes2015, shamrikova2022}.  Another concern is safety, as the process involves handling corrosive reagents and generating toxic waste that requires specialized disposal \cite{nelson1996total}.

Corresponding to N determination, the Kjeldahl method \cite{bremner1960determination} remains the global standard reference \cite{saez2013overview}. However, it demonstrates intrinsic disadvantages when faced with modern analytical demands. The procedure is notably time-consuming and labor-intensive, relying on multiple steps (digestion, distillation, and titration) that increase the probability of operational errors \cite{saez2013overview}. Similar to the Walkley-Black method \cite{jones1991kjeldahl}, serious criticism is directed at occupational and environmental safety due to the use of sulfuric acid, which generates hazardous waste. Additionally, the method lacks specificity as it quantifies total organic N, being unable to distinguish between protein and non-protein N without complex additional steps \cite{saez2013overview}. This creates a need for alternative methods that are faster, less destructive, and more environmentally sustainable, such as Near-Infrared (NIR) spectroscopy.

{To better understand the scientific foundations and relevance of this technique, it is important to examine the historical development of infrared radiation and the subsequent emergence of NIR spectroscopy as an analytical tool.} William Herschel, known for his experiment conducted on March 27, 1800, in which he used a prism to disperse sunlight beyond the visible spectrum, demonstrated the existence of radiation beyond the human eye's perception. Thus, infrared radiation and its region in the electromagnetic spectrum were discovered. However, the NIR region, which is contained within the infrared, was not used in scientific applications until approximately 1950; it was studied but considered to be of little use. It was only in the 1970s, with Karl Norris and his research group, that the NIR region began to be used for the analysis of agricultural samples (grains, seeds, wheat, soybeans), aiming to quantify chemical components (water, protein, lipids) \cite{BLANCO2002240}.

\begin{figure}[!ht]
\includegraphics[width=1.0\textwidth]{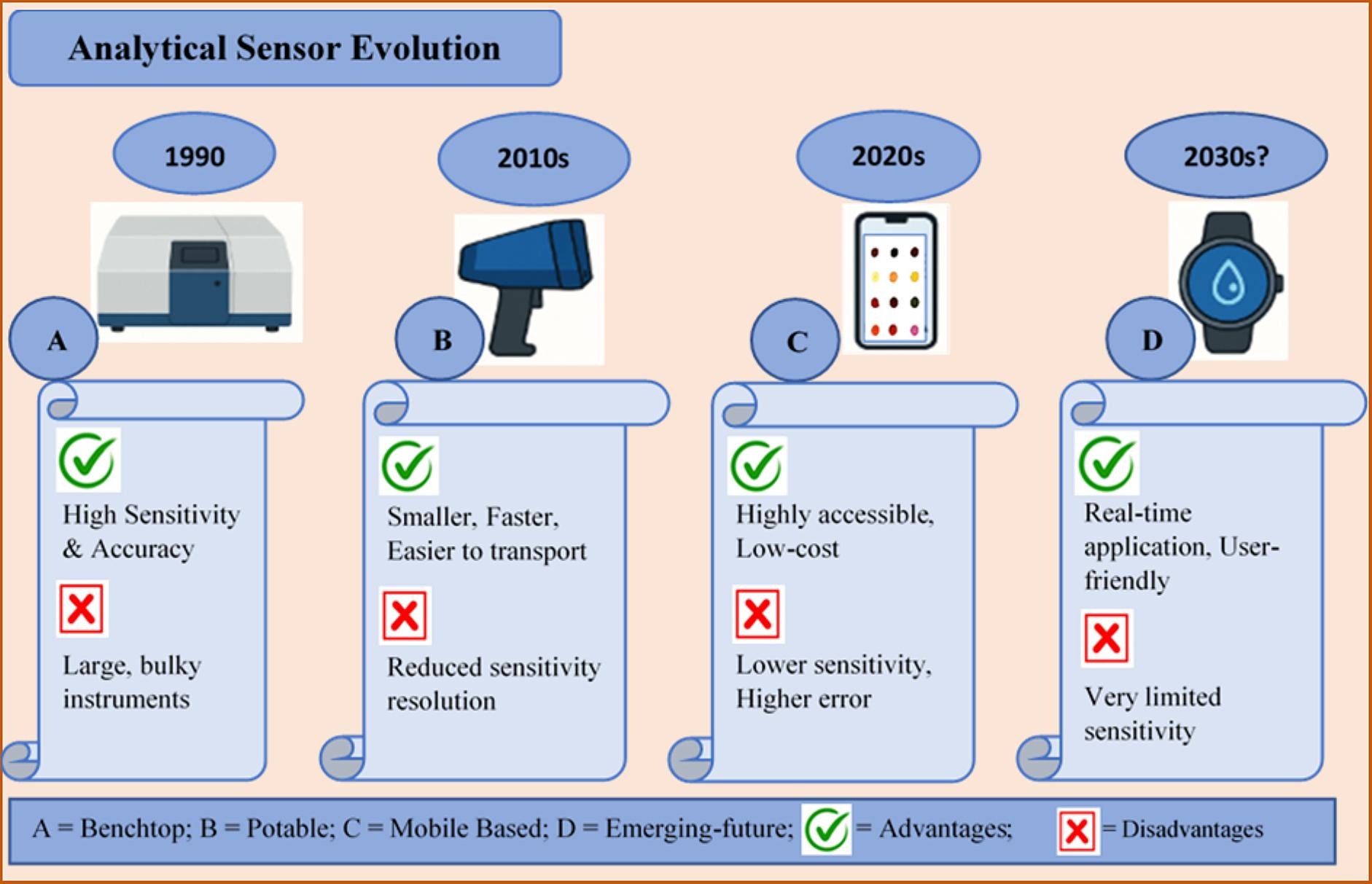}
\caption{Evolution of NIR analytical sensors \cite{ZAREEF2026126713}.}
\label{fig:tecnologiasnir}
\end{figure}

\begin{figure}[!ht]
\includegraphics[width=0.9\textwidth]{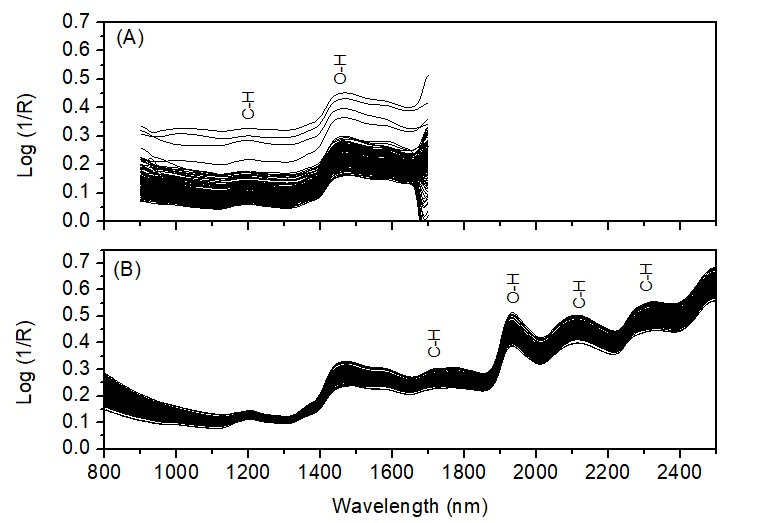}
\caption{Comparison between NIR spectra acquired by (A) portable equipment and (B) benchtop equipment \cite{https://doi.org/10.1002/cem.70073}.}
\label{fig:bancada-portatil}
\end{figure}

Figure \ref{fig:tecnologiasnir} illustrates the technological evolution of sensors for the NIR region over time, from the 1990s to the projection for the 2030s. The image compares four types of sensors: (A) Benchtop (1990s), which has high sensitivity and precision but is large and heavy; (B) Portable (2010s), which is small, fast, and easy to transport but does not have the same precision and sensitivity as benchtop models; (C) Mobile-based devices (2020s), which are extremely accessible and low-cost, but have much lower sensitivity and may increase error; and (D) Emerging and future wearable sensors (2030s), which are real-time, user-friendly technologies, but with even more reduced sensitivity. Each stage highlights the main advantages and disadvantages, demonstrating a trend towards greater accessibility and usability \cite{GOPAL2024117504}. {Nevertheless, the miniaturization and simplification of NIR devices can affect data quality and spectral resolution. These effects can be observed in Figure \ref{fig:bancada-portatil}, which compares spectra acquired using portable (A) and benchtop (B) equipment.}

It is observed that the spectra from the portable equipment have a considerably smaller spectral range, corresponding to approximately half the wavelength recorded by the benchtop equipment (Figure \ref{fig:bancada-portatil}). Another notable difference is the higher resolution and precision of the spectra obtained by the benchtop equipment. Additionally, the spectra from the portable equipment show distortions in the region near 1700 nm, particularly at the end of the spectral range.

In contrast to standard methods such as Walkley–Black \cite{jones1991kjeldahl} and Kjeldahl \cite{bremner1960determination} which require time-consuming sample preparation and analysis, NIR spectroscopy facilitates the simultaneous quantification of C and N in seconds, offering a rapid alternative. \citet{saez2013overview} underscore this operational delay as a major justification for developing novel analytical techniques.

The adoption of NIR spectroscopy is also justified by operational advantages. The non-destructive nature of the NIR technique is noteworthy, allowing for the physical preservation of the sample, enabling repeated readings or subsequent analyzes, as well as storage in soil banks \cite{viscarra2006determination}. Furthermore, the NIR technique offers a multiparametric capability, in which a single spectrum provides simultaneous information on various soil properties, eliminating the need for multiple chemical extraction procedures \cite{SORIANODISLA201724}.

From an economic perspective, although the initial investment in NIR instruments may be high, the operational cost per sample is drastically reduced due to the elimination of expensive chemical reagents and the reduction in required labor \cite{nocita2015soil, stenberg2010vis}. For example, according to studies by \cite{recena2019soil}, the \textit{Canadian Grain Commission} achieved savings of millions of dollars between 1973 and 1974 in protein analysis using NIR equipment, with an investment of only 96,000 Canadian dollars. The spectroscopic approach also helped reduce caustic waste by 47 t.

Commonly used for analyzing substances ranging from food and pharmaceuticals to soil and industrial compounds \cite{recena2019soil}, NIR spectroscopy operates based on molecular responses. Specifically, NIR spectroscopy measures energy absorptions derived from rotational and vibrational transitions within the molecules. The spectral region between 800 and 2500 nm (12500 to 4000 ~$cm^{-1}$) is predominantly characterized by overtone and combination absorption bands. Vibrational bands resulting from overtones, primarily related to organic compounds, appear in the 800 to 2000 nm range. Meanwhile, combination bands involving stretching and bending modes occur between 1800 and 2500 nm and are associated with specific chemical bonds \cite{workman2007practical}:

\begin{itemize}
    \item C=O (1900--2000 nm)
    
    \item C-H (1100--1225 nm, 1300--1420 nm, 1620--1800 nm, 2200--2460 nm)
    
    \item C-O, N-H (1400--1600 nm, 2000--2200 nm)
    
    \item O-H (1400--1600 nm, 1900--2000 nm, 2000--2200 nm)
\end{itemize}

Thus, the result of this technique is a complex spectrum, making visual analysis usually unfeasible for humans. In this context, the use of statistical tools becomes indispensable. This limitation can be mitigated by using machine learning (ML) to analyze these data, as demonstrated in \citet{wang2023assessment} and \citet{liu2023rapid} for C, and in \citet{sisouane2017prediction} for N.

According to \citet{bedin2021nir}, it is possible to create a regression curve that seeks to relate the data obtained from NIR spectroscopy of the samples to the actual N and C content. Therefore, to build predictors using ML methods, in addition to the spectra, the contents determined in the laboratory through bench methodologies such as the Walkley-Black and Kjeldahl methods \cite{jones1991kjeldahl} are necessary. 

It is hypothesized that portable NIR spectroscopy associated with ML techniques can generate reliable predictive models for the simultaneous quantification of C and N in soils, reducing analysis time and the dependence on conventional chemical methods. This study aimed to investigate the applicability of portable NIR spectroscopy combined with supervised, chemometric (PLS) or general-purpose (SVM, Ridge, Lasso and Random Forest), ML algorithms,  for the rapid and simultaneous quantification of C and N in different soil samples.

\section{Related Works}
\label{sec:related_works}

In recent years, the integration of NIR spectroscopy with data-driven modeling approaches has gained considerable attention for rapid soil characterization. This combination enables the development of fast, non-destructive, and cost-effective analytical strategies for predicting soil chemical and physical properties, supporting applications in precision agriculture and environmental monitoring. Several studies have demonstrated the potential of spectroscopic data combined with multivariate calibration and ML approaches to estimate soil attributes under different environmental and management conditions.

This section describes some relevant studies related to this work, as they use ML to estimate soil properties with regression models. Qualitative comparisons with these works and ours are discussed, focusing on the practical and theoretical aspects of their applications to real-world problems.

{Table \ref{tab:related_works} summarizes representative studies involving NIR spectroscopy combined with chemometric and ML algorithms for the prediction of different soil properties.}

\begin{sidewaystable}[!htp]
\caption{Related Works}
\label{tab:related_works}
\footnotesize
\begin{tabular}{ccccccc}
\hline
Reference &
\begin{tabular}[c]{@{}c@{}}Spectroscopic\\ Technique\end{tabular} &
  Spectral Range (nm) &
  Soil Dataset &
  Soil  Properties &
  Algorithms &
  \begin{tabular}[c]{@{}c@{}}Highest R²\end{tabular} \\ \hline
  
\cite{SORIANODISLA201724} &
\begin{tabular}[c]{@{}c@{}}vis-NIR and\\ MIR\end{tabular} &
  \begin{tabular}[c]{@{}c@{}}950–1650 and\\ 2500-25000\end{tabular} &
  \begin{tabular}[c]{@{}c@{}}Australian soils\\ (458 samples)\end{tabular} &
  \begin{tabular}[c]{@{}c@{}}RD, SAT, DUL,\\ OC, TC, TN,\\ EC, pH, CEC,\\ Ca2+, Mg2+, Na+,\\ K+, ESP, sand,\\ silt, and clay\end{tabular} &
  PLSR &
  \begin{tabular}[c]{@{}c@{}}0.79\end{tabular} \\ 

{\cite{sharififar2019}} &
NIR &
  \begin{tabular}[c]{@{}c@{}}1300–2500\\and 350–2500\end{tabular} &
  \begin{tabular}[c]{@{}c@{}}Australian soils\\ (151 samples)\end{tabular} &
  \begin{tabular}[c]{@{}c@{}}OC\\ TC\end{tabular} &
  \begin{tabular}[c]{@{}c@{}}SVR, PLSR, \\ and Cubist\end{tabular} &
  \begin{tabular}[c]{@{}c@{}}0.89\end{tabular} \\ 

{\cite{BENEDET2020114553}} &
\begin{tabular}[c]{@{}c@{}}pXRF\\ vis NIR DRS\end{tabular} &
  350–2500 &
  \begin{tabular}[c]{@{}c@{}}Brazilian soils\\ (315 samples)\end{tabular} &
  \begin{tabular}[c]{@{}c@{}}Sand, Silt,\\and Clay\end{tabular} &
  \begin{tabular}[c]{@{}c@{}}Gaussian,\\ SVM, and RF\end{tabular} &
  \begin{tabular}[c]{@{}c@{}}0.91\end{tabular} \\ 
  
{{\cite{TANG2020e00240}}} &
NIR &
  \begin{tabular}[c]{@{}c@{}}350–2500,\\ 1250–2500,\\and 900–1700  \end{tabular} &
  \begin{tabular}[c]{@{}c@{}}Australian Soil\\ (392 samples)\end{tabular} &
  \begin{tabular}[c]{@{}c@{}}pH, CEC, Ca,\\ Mg, clay, sand,\\ and TC\end{tabular} &
  \begin{tabular}[c]{@{}c@{}}Cubist\\ PLSR\end{tabular} &
  \begin{tabular}[c]{@{}c@{}}0.81\end{tabular} \\ 
  
{\cite{soilsystems6030066}} &
FT-NIR MEMS &
  1350-2500 &
  \begin{tabular}[c]{@{}c@{}}Italian Soils\\ (182 samples)\end{tabular} &
  \begin{tabular}[c]{@{}c@{}}SOC, sand,\\ silt, clay,\\ and CaCO3\end{tabular} &
  PLSR &
  \begin{tabular}[c]{@{}c@{}}0.83\end{tabular} \\ 
  
{{\cite{Dhawale2022Evaluation}}} &
Vis-NIR &
   \begin{tabular}[c]{@{}c@{}}5500–11000 \\ and 400–2220\end{tabular} &
  \begin{tabular}[c]{@{}c@{}}Canadian soils\\ (282 samples)\end{tabular} &
  Sand, clay, SOC &
  PLSR &
  \begin{tabular}[c]{@{}c@{}}0.82\end{tabular} \\ 

\cite{CARVALHO2022e00530} &
NIRS &
  1200–2400 &
  \begin{tabular}[c]{@{}c@{}}Southern Brazil\\ (2388 samples)\end{tabular} &
  SOM &
  \begin{tabular}[c]{@{}c@{}}MLR, PLSR, PCR,\\ SVM, RF, GPR\end{tabular} &
  \begin{tabular}[c]{@{}c@{}}0.70\end{tabular} \\ 

\cite{https://doi.org/10.1111/ejss.13323} &
MIRS &
  2500-25000 &
  \begin{tabular}[c]{@{}c@{}}Belgian soils\\ (375 samples)\end{tabular} &
  OC, P, K &
  \begin{tabular}[c]{@{}c@{}}SVM, GBR, RF, MLP,\\ Ensemble\end{tabular} &
  \begin{tabular}[c]{@{}c@{}}0.88\end{tabular} \\ 

\cite{SABERIOON2024109494} &
Vis–NIR–SWIR &
  350–2500 &
  \begin{tabular}[c]{@{}c@{}}LUCAS database\\ (11441 samples)\end{tabular} &
  SOC &
  \begin{tabular}[c]{@{}c@{}}1DCNN, FCNN,\\ SAE, RF\end{tabular} &
  \begin{tabular}[c]{@{}c@{}}0.78\end{tabular} \\ 

\cite{https://doi.org/10.1155/2019/3563761} &
VIS-NIR &
  400–2450 &
  \begin{tabular}[c]{@{}c@{}}Chinese soils\\ (248 samples)\end{tabular} &
  SOM &
  \begin{tabular}[c]{@{}c@{}}BPN, MLP,\\ LeNet5, DenseNet10\end{tabular} &
  \begin{tabular}[c]{@{}c@{}}0.89\end{tabular} \\ 

\textbf{Ours} &
NIR &
  900-1700 &
  \begin{tabular}[c]{@{}c@{}}Brazillian Soils\\ (153 samples)\end{tabular} &
  C, N &
  \begin{tabular}[c]{@{}c@{}}PLS, SVR, Ridge,\\ RF, Ensemble\end{tabular} &
  \begin{tabular}[c]{@{}c@{}}0.91\end{tabular} \\ \hline
\end{tabular}
\end{sidewaystable}

{Overall, the reviewed studies demonstrate the strong potential of NIR spectroscopy combined with chemometric and ML algorithms for rapid and non-destructive prediction of soil properties. In general, the studies presented corroborate the review conducted by \citet{gozukara2025prediction}, which evaluated 305 published studies involving Vis–NIR, MIR, and pXRF spectroscopy associated with chemometric and ML approaches for the prediction of several soil physical and chemical properties.

This work presents significant differentiators compared to previous studies in four main aspects: (1) sample diversity, with the systematic inclusion of two distinct soil types (Oxisol and Inceptisol); (2) spectroscopic acquisition methodology, using the \textit{MyNIR} equipment (900 - 1700 nm); (3) multiple preprocessing algorithms with automated hyperparameter optimization; (4) an innovative Ensemble Learning architecture that integrates the complementary advantages of PLS, Ridge, SVR, and RF to maximize the predictive power of the final model and minimize the effects of overfitting and underfitting; (5) comparison of validation strategies: cross-validation versus the holdout method, as well as optimal split proportions for the train and test sets; and (6) an insightful analysis of feature importance, revealing the most and least important wavelengths of the portable NIR device's spectrum for prediction using ML models.

\section{Material and Methods} \label{sec:maerial_and_methods}

The Materials and Methods section was structured in a logical and chronological order to clearly describe the experimental design and the methodological workflow adopted throughout the study. The section is organized as follows: (1) Sample collection and preparation; (2) NIR spectral acquisition; (3) Reference analysis of carbon and nitrogen; (4) Exploratory Data analysis; (5) Data preprocessing; (6) Logarithmic transformation applied to NIR reflectance data; (7) adaptative Spectral Trimming; (8) Mean Centering; (9) Savitzky-Golay Filter; (10) Removal of samples outside the reflectance pattern; (11) robust Outlier Removal based on Huber Loss; (12) Outlier removal methods comparison; (13)  comparison between validation strategies; (14) egression model building; (15) Model evaluation. This methodological organization was designed to ensure clarity, reproducibility, and a comprehensive understanding of the experimental sequence and analytical procedures performed in the study. Figure \ref{fig:fluxoingles} shows more details.

\begin{figure}[!ht] 
\centering 
\includegraphics[width=0.35\textwidth]{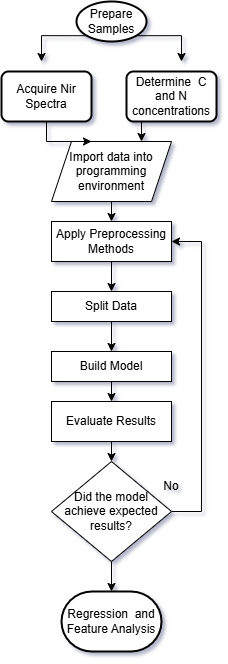} 
\caption{Simplified Fluxogram describing the methodology used in this study} 
\label{fig:fluxoingles} 
\end{figure}

\subsection{Sample Collection and Preparation}
\label{subsec:data_acquisition}
This work analyzes two soils: Inceptisol and Oxisol, {aiming to evaluate the influence of preprocessing methods and ML models on the prediction performance of soil attributes using NIR spectroscopy. A total of  153 soil samples, 72 Oxisol and 81 Inceptisol, were collected from different locations and management conditions. The soil samples were air-dried at room temperature, homogenized, and sieved through a 2 mm mesh prior to analysis.}

\subsection{Reference Analysis of Carbon and Nitrogen}

Total C was determined using the Walkley–Black method \cite{walkley1934examination}, a volumetric procedure based on potassium dichromate (K$_2$Cr$_2$O$_7$) oxiddation. Briefly, soil organic C was oxidized to carbon dioxide (CO$_2$), under acidic conditions, while chromium (Cr) was reduced from Cr(VI) to Cr (III). The excess dichromate remaining after oxidation has to be back-titrated with a standardized ammonium ferrous sulfate solution. The amount of dichromate consumed during the reaction is proportional to the organic carbon content of the soil sample, allowing its quantification \cite{da2009manual, walkley1934examination}.

Total N was determined according to the Kjeldahl method described by \cite{jones1991kjeldahl} and adapted by \cite{tedesco1995}, with minor modifications to improve analytical sensitivity and accuracy. In summary, dried and sieved soil samples were subjected to acid digestion in the presence of a catalytic mixture,  converting organic N into ammonium ions NH4+. After the extraction, the extracts were cooled, diluted with deionized water, and alkalized with sodium hydroxide (NaOH), promoting the conversion of  ammonium to ammonia (NH$_3$). The realeased ammonia was distilled and trapped in a boric acid (H$_3$BO$_3$) solution, followed by quantification to determine the total N of the soil samples.

\subsection{NIR Spectral Acquisition}

{The spectral acquisition of samples was performed using one NIR instrument, the portable device myNIR model from spectral Solutions (China).
In the MyNIR spectrometer, 20 cm$^{3}$ of the sample was added to the Petri dish, and all spectra were acquired in the reflectance mode with less rotation over
the range from 11,104 to 5,881 cm$^{-1}$ (equivalent to 900 nm - 1700 nm) with a
resolution of 6.1 cm$^{-1}$ using the Spectral Solutions software (Spectral Solutions,
China). Each spectrum had an average of 150 scans. Each sample was acquired
twice. A schematic view of the measurement system used in this study is shown
in Figure \ref{fig:nir_acq}. This system consists of a portable NIR spectrometer that connects to a desktop computer via a USB cable.}

\begin{figure}[!ht]
    \centering
    \includegraphics[width=140mm]{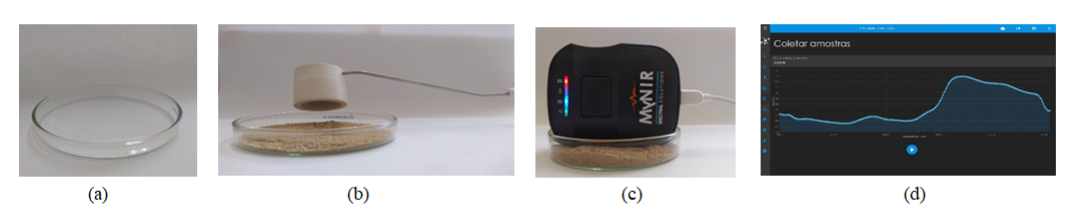}
    \caption{Illustrative representation of the NIR spectra acquisition using a portable instrument: (A) Petri dish, (b) Petri dish containing 20 cm$^3$ of sample, (c) MyNIR spectrometer, and (d) spectrum obtained by software \cite{https://doi.org/10.1002/cem.70073}.}
    \label{fig:nir_acq}
\end{figure}
Defesa de

\subsection{Exploratory Data Analysis}\label{subsec:EDA}

Before applying preprocessing methods and fitting ML models, as a means to provide an exploratory baseline to guide the selection of optimal data treatment strategies, an exploratory data analysis (EDA) was performed. EDA is a fundamental step in this study for it shows, visually and numerically, where the problems or critical information about the datasets may lie.  The data analyzed consist of: (i) The independent variable, which consists of 153 NIR spectra collected in duplicate, for each type of soil, resulting in a $162x256$ matrix represented as $X_i$ (Inceptisol spectra) and another $144x256$ matrix represented as $X_o$ (Oxisol spectra); (ii) The dependent variable y, that represents the total concentration (in percentage) of C and N of each sample, resulting in two vertical vectors of $144x1$ for the Oxisol soil type and two vertical vectors of $162x1$ for the Inceptisol soil type.

Corresponding to the independent variable $X$ (Features), which is the NIR spectrum of all samples combined, the EDA reveals the exact wavelengths of the features, the number of features, and the reflectance values. Figures \ref{fig:inceptisol_show} and \ref{fig:oxisol_show} show the reflectance values of samples versus wavelengths.

\begin{figure}[!h]
\includegraphics[width=1.0\textwidth]{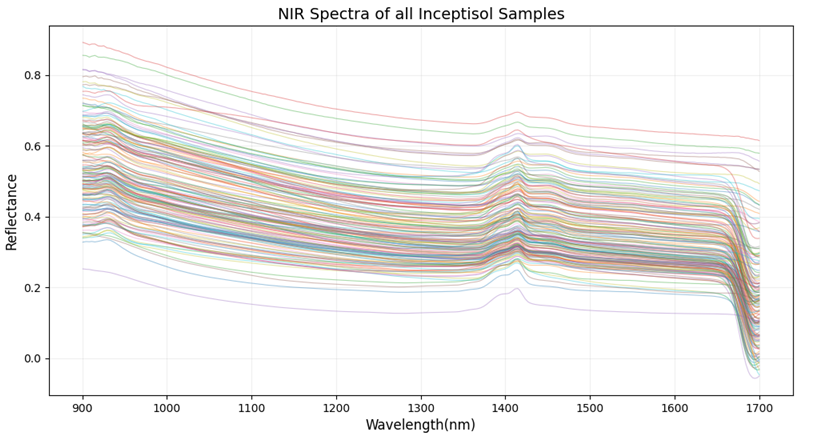}
\centering
\caption{Inceptisol Spectra}
\label{fig:inceptisol_show}
\end{figure}

\begin{figure}[!h]
\includegraphics[width=1.0\textwidth]{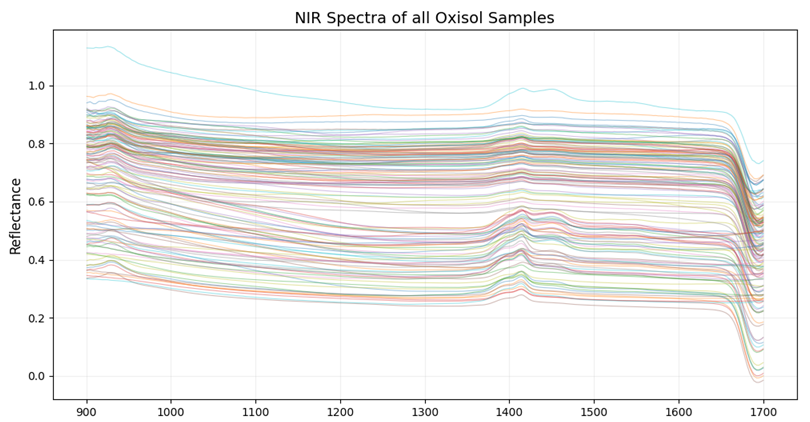}
\centering
\caption{Oxisol Spectra}
\label{fig:oxisol_show}
\end{figure}

Both plots suggest that there are some instrumental artifacts, such as distortion, at the end of the spectra mentioned.

Tables \ref{tab:inceptisol_summary} and \ref{tab:_oxisol_summary} provide numerical details regarding the spectra of both types of soil samples, such as the maximum, minimum, and mean values of NIR reflectance, as well as the wavelength interval of the portable device. Dimensions (shape) are also included, corresponding to the matrix in the programming environment.

\begin{table}[!ht]
\centering
\caption{Structure and Statistical Summary of Spectral Data (X) from Inceptisol Samples}
\label{tab:inceptisol_summary}
\begin{tabular}{p{4cm} p{3.5cm} p{3.5cm}}\hline
\textbf{Category} & \textbf{Property} & \textbf{Value / Range} \\ \hline
\textit{Matrix Structure} & Dimensions (Shape) & (162, 256) \\
                          & Spectral Features  & 256 variables \\ [0.5cm]
\textit{Wavelength (wl)}  & Range              & 900.56 -- 1700.16 \\ [0.5cm]
\textit{Reflectance (X)}  & Minimum            & -0.0565 \\
                          & Maximum            & 0.8923 \\
                          & Mean               & 0.3901 \\ \hline
\end{tabular}
\end{table}

\begin{table}[!ht]
\centering
\caption{Structure and Statistical Summary of Spectral Data (X) from Oxisol Samples}
\label{tab:_oxisol_summary}
\begin{tabular}{p{4cm} p{3.5cm} p{3.5cm}}
\hline
\textbf{Category} & \textbf{Property} & \textbf{Value / Range} \\ 
\midrule
\textit{Matrix Structure} & Dimensions (Shape) & (144, 256) \\
                          & Spectral Features  & 256 variables \\[0.5cm]

\textit{Wavelength (nm)}  & Range              & 900.56 -- 1700.16 \\[0.5cm]

\textit{Reflectance (X)}  & Minimum            & -0.0230 \\
                          & Maximum            & 1.1346 \\
                          & Mean               & 0.6141 \\ \hline
\end{tabular}
\end{table}

The maximum and minimum properties show that there are values not contained in the theoretical physical interval range of 0 to 1 in the reflectance metric. Thus, indicating either noise, human error in collecting the spectra, or even equipment malfunction.

A brief analysis of the datasets for the independent variable $y$ (target) is performed for 72 Oxisol samples and 81 Inceptisol samples, as shown in Tables \ref{tab:cambistats} and \ref{tab:latostats}.

\begin{table}[!ht]
\centering
\caption{Statistics of Inceptisol samples: C and N concentrations}
\begin{tabular}{p{4cm}p{3cm}p{3cm}}\hline
 & \textbf{C (\%)} & \textbf{N (\%)} \\ \hline
Minimum & 0.17 & 0.02 \\
Maximum  & 6.37 & 0.48  \\
Mean & 2.35 & 0.17 \\
Standard Deviation & 1.39 & 0.10 \\ \hline
\end{tabular}
\label{tab:cambistats}
\end{table}

In Inceptisol, C concentrations ranged between 0.17 and 6.37\%, with a mean of 2.35\% and a standard deviation (SD) of 1.39\%, while N concentrations varied between 0.02 and 0.48\%, with a mean of 0.17\% and a SD of 0.10\%.

\begin{table}[!ht]
\centering
\caption{Statistics of Oxisol samples: C and N concentrations}
\begin{tabular}{p{4cm}p{3cm}p{3cm}}\hline
 & \textbf{C (\%)} & \textbf{N (\%)} \\ \hline
Minimum & 0.33 & 0.04 \\
Maximum & 6.62 & 0.56 \\
Mean & 2.74 & 0.21 \\
Standard Deviation & 1.50 & 0.13 \\ \hline
\end{tabular}
\label{tab:latostats}
\end{table}

In Oxisol, C concentrations are observed to be slightly higher, with a minimum of 0.33\%, a maximum of 6.62\%, a mean of 2.74\%, and a SD of 1.50\%. Similarly, N concentrations in Oxisol are also higher, varying between 0.04 and 0.56\%, with a mean of 0.21\% and a SD of 0.13\%.

The data used in this study are available in the public repository \url{https://github.com/ViniWiee/NIR-datasets}, allowing replication of the study.

\subsection{Data preprocessing}
\label{subsec:data_preprocessing}

Several factors can affect the quality of infrared spectra. A well-defined protocol for sample preparation and analysis is necessary for each spectrometer. Aiming to reduce these factors and thereby improve the predictive capability of the models, it is strongly recommended to apply preprocessing to the raw data \cite{barra2021soil}.

The preprocessing stage is fundamental to the success of ML applications. Examples of preprocessing methods include filters for noise removal, signal smoothing, removal of undesirable samples, and data normalization. \cite{sharma2022study}. Table \ref{quad:resumo_preprocessamento} briefly presents all preprocessing methods utilized in this study, as well as a short description of each.

\begin{table}[!htbp]
\centering
\footnotesize
\caption{Summary of preprocessing methods applied to spectral data.}
\label{quad:resumo_preprocessamento}
\begin{tabular}{lcc}\hline
\textbf{Method} & \textbf{Objective} & \textbf{Parameters/Settings} \\
\midrule
MSC & \begin{tabular}[c]{@{}c@{}} Correction of light \\scattering effects\end{tabular} & \begin{tabular}[c]{@{}c@{}}Mean spectrum as\\ reference \end{tabular} \\ 

SNV + Detrending & \begin{tabular}[c]{@{}c@{}}  Scatter correction without\\ reference \end{tabular} & ---  \\

Logarithmic Transf. & \begin{tabular}[c]{@{}c@{}} Linearization of the\\ absorbance-reflectance\\ relationship\end{tabular} & ---\\

Spectral Trimming & \begin{tabular}[c]{@{}c@{}} Elimination of regions with \\low signal-to-noise ratio\end{tabular} & \begin{tabular}[c]{@{}c@{}}Based on PLS \\performance \end{tabular} \\

Mean Centering & \begin{tabular}[c]{@{}c@{}} Centering variables around\\ the mean\end{tabular} & --- \\

Savitzky-Golay & \begin{tabular}[c]{@{}c@{}} Spectral smoothing and\\ noise reduction\end{tabular} & \begin{tabular}[c]{@{}c@{}}$w, p, d$ optimized by grid\\ search \end{tabular} \\

Outlier Removal & \begin{tabular}[c]{@{}c@{}} Identification of samples\\ with high prediction error\end{tabular} & \begin{tabular}[c]{@{}c@{}}PLS with Huber Loss\\ ($2.5\sigma$ threshold) \end{tabular} \\

Standardization & \begin{tabular}[c]{@{}c@{}} Scaling to zero mean and\\ unit variance\end{tabular} & --- \\ 
\bottomrule
\end{tabular}
\end{table}

\subsubsection{\textit{Multiplicative Scatter Correction} - MSC}\label{subsubsec:msc}

The motivation for applying MSC lies in a recurring empirical observation: when the spectral values of an individual sample are plotted against the mean spectral values of a calibration set, the resulting points tend to be linearly distributed. It was found that different samples generate lines with characteristic slopes and offsets (intercepts). These systematic variations are predominantly attributed to physical light scattering effects, which alter the baseline and scale of the spectrum in an additive and multiplicative manner, respectively. Therefore, given this linear relationship, the MSC procedure is applied to standardize the spectra. The method identifies and corrects the unique slope and offset of each sample, aligning them to a common reference. The result is the attenuation of scattering effects, thereby enhancing the information related to the chemical composition of the samples \cite{isaksson1988effect}.

The method's operation is based on the linear model in Equation \ref{eq:msc1}, where $x_i$ is the spectrum of sample $i$, $\bar{x}$ is the mean calibration spectrum, $a_i$ is the additive scattering effect (intercept), $b_i$ is the multiplicative scattering effect (slope), and $e_i$ represents the residuals. The coefficients $a_i$ and $b_i$ are calculated by linear regression according to Equations \ref{eq:msc2} and \ref{eq:msc3}, where $x{i,k}$ is the spectral value of sample $i$ at wavelength $k$, $\bar{x}_k$ is the mean value at wavelength $k$, $\bar{x}_i$ is the mean of the values for sample $i$, and $\bar{x}$ is the overall mean of all spectra.

\begin{equation}\label{eq:msc1}
x_i = a_i + b_i \bar{x} + e_i
\end{equation}

\begin{equation}\label{eq:msc2}
b_i = \frac{\sum_{k=1}^{K} (x_{i,k} - \bar{x}_i)(\bar{x}k - \bar{x})}{\sum{k=1}^{K} (\bar{x}_k - \bar{x})^2}
\end{equation}

\begin{equation}\label{eq:msc3}
a_i = \bar{x}i - b_i \bar{x}
\end{equation}

Finally, the final correction is applied by the transformation in Equation \ref{eq:msc4}, which removes the additive effect ($x_i - a_i$), corrects the multiplicative effect ($b_i$), and results in the corrected and standardized spectrum ($x_{i}^{\text{corr}}$).

\begin{equation}\label{eq:msc4}
x_{i}^{\text{corr}} = \frac{x_i - a_i}{b_i}
\end{equation}

\subsubsection{\textit{Standard Normal Variate} and \textit{De-trending} (SNV-DT)}\label{subsubsec:snv}

The sequential application of SNV and De-trending (DT) techniques was proposed to correct the effects of light scattering in diffuse reflectance spectra \cite{barnes1989standard}. The SNV method aims to correct the multiplicative and additive effects arising from variations in particle size and optical path, normalizing each spectrum individually by its mean and SD. Subsequently, the DT method removes residual polynomial trends, whether linear or quadratic, in the spectra that result from the wavelength dependence of the scattering coefficient. The combined application of these two methods linearizes the relationship between the spectra and the analytical concentration, significantly improving the performance of calibration models.

Thus, for the SNV method, let a spectrum $i$ be represented by a vector of absorbances or reflectances $x_i = [x_{i1}, x_{i2}, ..., x_{ip}]$, where $p$ is the number of variables (wavelengths). The value transformed by SNV for wavelength $j$ is given by Equation \ref{eq:snv1}, where each spectral point $x_{ij}$ is centered by the mean $\bar{x}_i$, which is calculated according to Equation \ref{eq:snv2}, and $s_i$ is the standard deviation of all points in the spectrum, calculated by Equation \ref{eq:snv3}.

\begin{equation}\label{eq:snv1}
x_{ij(\text{snv})} = \frac{x_{ij} - \bar{x}_i}{s_i}
\end{equation}

\begin{equation}\label{eq:snv2}
\bar{x}i = \frac{1}{p} \sum{j=1}^{p} x_{ij}
\end{equation}

\begin{equation}\label{eq:snv3}
s_i = \sqrt{\frac{\sum_{j=1}^{p} (x_{ij} - \bar{x}_i)^2}{p - 1}}
\end{equation}

Next, a polynomial of order $m$ is fitted to the SNV-corrected spectrum and then subtracted to remove the curvilinear trend (De-Trending). The fitted polynomial $\hat{t}_{ij}$ for wavelength $j$ in spectrum $i$ is defined by Equation \ref{eq:snv4}, where $\lambda_j$ is the wavelength of index $j$ and $a_0, a_1, ..., a_m$ are the coefficients of the polynomial, determined by a least squares regression of the SNV spectrum $x{ij(\text{snv})}$ against the wavelengths $\lambda_j$.

\begin{equation}\label{eq:snv4}
\hat{t}{ij} = a_0 + a_1 \lambda_j + a_2 \lambda_j^2 + \dots + a_m \lambda_j^m
\end{equation}

The final value (SNV - DT) is obtained by subtracting the fitted trend according to Equation \ref{eq:snv5}

\begin{equation}\label{eq:snv5}
x_{ij(\text{snv-dt})} = x_{ij(\text{snv})} - \hat{t}_{ij}
\end{equation}

\subsection{Logarithmic transformation applied to NIR reflectance data}\label{subsec:log}

The logarithmic transformation was applied to the NIR reflectance data to linearize the relationship between absorbance and reflectance. This approach is based on the Beer-Lambert principle, which explains how light is absorbed as it passes through or is reflected by matter. This principle states that the higher the concentration ($c$) of a chemical in a liquid environment, the higher is the absorption ($a$) of light that is irradiated through that compound. Briefly, $c$ and $a$ are directly proportional, allowing the determination of the chemical concentration in a liquid with light alone \cite{goetz1985imaging}.

The Total light interaction in a system i, by the Conservation of Energy Equation: $Reflected + Absorbed + Transmitted = 100\%$ \cite{skoog2017principles}. Thus, it is possible to correlate diffuse reflectance (which is the default NIR metric) to absorbed. This transformation was implemented through the logarithmic function $\log(1/R)$ as $R$ represents the reflectance, thus converting the data into a pseudo-linear absorbance scale.

\subsection{Adaptative Spectral Trimming}\label{subsec:trim}

Several preprocessing methods can be combined for the spectral cleaning/\\adjustment. One of these techniques is the removal of visibly noisy regions at the edges of the spectrum. Based on the premise that regions at the spectrum edges can contain significant noise that compromises predictive modeling, this study developed a systematic method for identifying optimal spectral intervals \cite{10.1371/journal.pone.0105708}.  

The technique proposed in this study operates through an iterative process of refining the spectral interval using the PLS model. The method starts with the full spectrum and, in each iteration, sequentially removes one feature from the extremities of the spectral domain, trains a new PLS model, and evaluates its performance using the coefficient of determination R². At each step, the algorithm compares the obtained R² with the historical best value, storing the spectral configuration that produced the highest predictive power. This process continues until the predefined minimum number of features is reached, ensuring that a sufficient amount of spectral information is maintained for training.

By replacing the manual and visual aspects with this data-driven approach, the feature selection process becomes statistically verifiable and iterative. This methodology eliminates the subjectivity inherent in manual spectral trimming, ensuring that the optimal interval is determined by measurable performance metrics rather than researcher bias. While particularly robust for datasets with high sample homogeneity, the algorithm remains highly effective across diverse data types. Ultimately, this method functions as a performance-oriented adaptive filter, systematically isolating the most informative spectral regions to enhance model accuracy.

\subsection{Mean Centering}\label{subsec:padronizacao}

The mean centering (MC) is a widely used data preprocessing method, consisting of centering the data around zero by removing the mean of each variable, thus emphasizing the differences between samples\cite{rinnan2009review}

According to \citet{wold1987principal}, MC is an essential step in PLS analysis and principal component analysis (PCA) for spectroscopic data. By subtracting the mean from each variable, the vertical offset that can be caused by light scattering, differences in reflectance, or other physical interferences is removed. The model is thus driven to capture the variations around the mean, which contain relevant chemical information.

Given a matrix of spectroscopic data represented by $\mathbf{X}$ with $n$ samples and $p$ variables (wavelengths), as presented in Equation \ref{eq:mc1}.

\begin{equation}\label{eq:mc1}
\mathbf{X} =
\begin{bmatrix}
x_{11} & x_{12} & \cdots & x_{1p} \\
x_{21} & x_{22} & \cdots & x_{2p} \\
\vdots & \vdots & \ddots & \vdots \\
x_{n1} & x_{n2} & \cdots & x_{np}
\end{bmatrix}
\end{equation}

The mean for each variable $j$ is given by Equation \ref{eq:mc2}, and the mean-centered matrix $\mathbf{X}_c$ is obtained using Equation \ref{eq:mc3}.

\begin{equation}\label{eq:mc2}
\bar{x}j = \frac{1}{n} \sum{i=1}^{n} x_{ij}, \quad \text{for } j = 1, 2, \ldots, p
\end{equation}

\begin{equation}\label{eq:mc3}
\mathbf{X}_c = \mathbf{X} - \mathbf{1}\bar{\mathbf{x}}^T
\end{equation}
where $\mathbf{1}$ is a column vector of ones and $\bar{\mathbf{x}} = [\bar{x}_1, \bar{x}_2, \ldots, \bar{x}_p]^T$.

\subsection{Savitzky-Golay Filter}\label{subsec:reffiltro}

Common treatments in vibrational spectroscopy include smoothing to eliminate high-frequency noise and derivatives, which reduce baseline drift and emphasize specific spectral features \cite{barra2021soil}.

Smoothing functions as a low-pass filter designed to suppress high-frequency noise. This operation is performed independently on each row of the data matrix, targeting adjacent variables. The technique is based on the principle that neighboring variables (\textit{i.e.}, adjacent columns) are highly correlated and contain similar information. By averaging these values, noise is reduced without significant loss of the underlying signal of interest \cite{savitzky1964smoothing}.

Derivatives are widely used to remove irrelevant baseline signals by calculating the derivative of the spectral response with respect to the variable index or another relevant scale, such as wavelength or wavenumber. Functioning as a high-pass filter with frequency-dependent scaling, derivatives are particularly effective when broad, low-frequency features (e.g., baselines) interfere with the narrow, high-frequency features that contain the primary signal of interest \cite{savitzky1964smoothing}.

Both smoothing and differentiation are typically integrated into a single operation via the SG filter. As established by \citet{savitzky1964smoothing}, this approach relies on fitting a low-degree polynomial to a sliding window of adjacent data points through linear least squares. The filter’s strength lies in its ability to extract derivatives of various orders directly from the polynomial coefficients. This allows for noise attenuation and the resolution of overlapping spectral bands without the significant peak distortion or 'flattening' common in simpler moving-average filters. By leveraging this unified framework, the signal-to-noise ratio is improved while preserving the essential information of the spectra.

\subsection{Removal of samples outside the reflectance pattern}\label{subsec:reflectance_outliers}

The MyNIR spectrometer provides reflectance spectra in which feature values range from 0 to 1, representing the proportion of light reflected by the sample at each wavelength.

To ensure data integrity, a systematic filtering procedure was implemented to identify and remove samples with reflectance values outside this physical range. The algorithm performs a point-by-point validation of each spectral signature in matrix $\mathbf{X}$. Any sample containing at least one value falling outside the [0, 1] interval is flagged as an outlier and excluded from subsequent modeling stages. Finally, the procedure generates a quantitative report summarizing the total count of validated samples, rejected entries, and the specific number of inconsistent values identified across the dataset.

\subsection{Robust Outlier Removal based on Huber Loss}\label{subsec:reflectance_outliers}

Following the approach by \citet{bjerrum2017data}, we coupled NIPALS with Huber-based weighting to identify and remove samples associated with disproportionately high prediction errors, thereby enhancing model robustness, refining the training set and improving model generalization.

This junction integrates the advantages of both quadratic and linear estimators. Unlike the Mean Squared Error (MSE), which is highly sensitive to outliers, the Huber loss behaves quadratically for small residuals and linearly for larger ones. This dual behavior is defined in Equation \ref{eq:huberloss}, where $\delta$ represents the threshold parameter that governs the transition between the two regimes, and $a$ denotes the prediction error (residual).

\begin{equation}\label{eq:huberloss}L_\delta(a) =\begin{cases}\frac{1}{2}a^2 & \text{for } |a| \leq \delta, \\delta\left(|a| - \frac{1}{2}\delta\right) & \text{for } |a| > \delta,\end{cases}\end{equation} 

The Nonlinear Iterative Partial Least Squares (NIPALS) algorithm is a standard procedure for performing PCA and PLS through iterative decomposition. As described by \citet{GELADI19861}, NIPALS factorizes the predictor ($\mathbf{X}$) and response ($\mathbf{Y}$) matrices into latent components (scores and loadings) to maximize their mutual covariance. For a mean-centered matrix $\mathbf{X} \in \mathbb{R}^{n \times m}$, the algorithm iteratively computes the score vectors $\mathbf{t}_h$ and loading vectors $\mathbf{p}_h$ for each component $h$. This refinement continues until the difference between iterations falls below a tolerance $\epsilon$ (typically $10^{-6}$) or reaches a maximum $N$. Following each extraction, the matrix is deflated ($\mathbf{X} \leftarrow \mathbf{X} - \mathbf{t}_h \mathbf{p}_h^\top$) to ensure the orthogonality of subsequent components.

\subsection{Outlier removal methods comparison}

Comparing the two outlier removal strategies reveals distinct behaviors influenced by soil classification and the specific chemical constituents analyzed. The evaluation follows the methodologies previously established in this study: (i) a physical threshold filtering based on reflectance limits, and (ii) a robust statistical refinement using NIPALS-PLS regression coupled with Huber loss residual analysis.

For the Inceptisol dataset, the physical filter targeted spectrally anomalous samples—those with reflectance values outside the $[0, 1]$ interval. This criterion led to the exclusion of 14 samples (8.64\% of the original 162-sample set). Notably, all excluded samples exhibited negative reflectance values, with a minimum recorded value of ${-}$0.05, while no values exceeded the upper limit. These inconsistencies are likely attributable to instrumental noise or transient failures during spectral acquisition.

Regarding the high-residual method, samples with prediction errors exceeding 2.5 SD of the residuals were classified as outliers. This approach removed 13 samples from the C model and 12 from the N model. Interestingly, no overlap was observed between the samples flagged by physical filtering and those identified by the statistical residual method.

In the Oxisol dataset, the physical filter resulted in the removal of only four samples; three exhibited negative reflectance, while one exceeded the physical limit with a value of 1.13. Compared to the Inceptisol group, the Oxisol data showed fewer physically inconsistent samples and fewer extreme outliers, suggesting higher stability during acquisition. Statistical refinement identified 22 outliers in the C model and 19 in the N model. Again, no overlap between methods was found. However, the total number of statistically removed samples was higher for Oxisols than for Inceptisols, indicating greater multivariate variability in the Oxisol spectra. Furthermore, the C models demonstrated higher sensitivity to outliers than the N models.

Detailed indices and specific sample characteristics for both methods are provided in Tables \ref{tab:high_prediction} and \ref{tab:out_of}. Lower indices (\textit{e.g.}, 7 and 19) correspond to the initial spectral region, whereas higher values represent the mid-to-late regions of the spectrum. Since most of the indices for the C and N model datasets are identical, no distinction has been made in the tables below.

\begin{table}[h]
\centering
\caption{Samples with high prediction error by soil type.}
\label{tab:high_prediction}
\begin{tabular}{lp{10cm}}
\hline
\textbf{Soil} & \textbf{High Prediction Error Samples (index)} \\ \midrule
\textbf{Inceptisol} & 5, 15, 40, 43, 46, 47, 51, 66, 70, 72, 85, 111, 152, 153 \\ \addlinespace
\textbf{Oxisol}     & 0, 7, 24, 25, 29, 30, 36, 56, 60, 64, 78, 80, 81, 82, 83, 84, 93, 95, 96, 98, 99, 129 \\ \bottomrule
\end{tabular}
\end{table}

\begin{table}[h]
\centering
\caption{Samples out of interval by soil type.}
\label{tab:out_of}
\begin{tabular}{lp{10cm}}
\hline
\textbf{Soil} & \textbf{Out of Interval Samples (index)} \\ \midrule
\textbf{Inceptisol} & 58, 59, 108, 109, 110, 112, 114, 115, 117, 128, 129, 130, 131, 140 \\ \addlinespace
\textbf{Oxisol}     & 19, 115, 121, 133 \\ \bottomrule
\end{tabular}
\end{table}

\subsection{Comparison between validation strategies}
\label{subsec:training-test-sets}

The first step in building a predictive model is to divide the spectrum into training and testing sets (the Holdout method). Dividing the data into training and testing sets allows us to check if the model performs well not only for its fitting (training) but also for its generalization capability (testing). The most common splits are 60:40, 70:30, and 80:20 \cite{raschka2019python}. 

Several data splitting methods have been proposed in the literature \cite{ferreira2021kennard}. Typically, data are split randomly, and unfortunately, this may not be the most efficient strategy for small datasets. In this work, the Kennard-Stone algorithm was used in Python via the Astartes package to split the data into training and testing sets. The method proposed by \citet{kennard1969computer} is widely used in chemometrics for improving results, as it being fast and easy to use and does not rely on randomness. The method consists of capturing the greatest possible variability from the original set to create models that exhibit statistically significant differences in order to improve predictive capability. For this, the Euclidean distance is used for each pair of samples, with the most distant ones being selected to compose the training set, as presented in Equation \ref{eq:dist_euclidiana}, where:

\begin{itemize}
\item $d(\mathbf{p}, \mathbf{q})$: represents the Euclidean distance between the feature vectors of samples $\mathbf{p}$ and $\mathbf{q}$;
\item $x_p(i)$ and $x_q(i)$: denote the values of the $i$-th feature for the respective samples;
\item $j$: is the dimensionality of the space (the total number of features);
\item $N$: is the total number of samples in the original set.
\end{itemize}

\begin{equation}\label{eq:dist_euclidiana}
dx(\mathbf{p}, \mathbf{q}) = \sqrt{\sum_{i=1}^{j} (x_p(i) - x_q(i))^2}\ p, q\in [1, N]
\end{equation}
in which:

The metric quantifies the geometric dissimilarity between two samples in the $j$-dimensional space, serving as the basis for the algorithm's selection strategy, which prioritizes the greatest sample dispersion to optimize the representativeness of the training set.

However, this approach might face challenges, such as choosing the best proportion for the split. Correspondingly, the 10-fold cross-validation and the leave-one-out strategy were used to better estimate the MSE.

K-fold cross-validation partitions the data into $k$ equal (or approximately equal) parts randomly to train the model and select hyperparameters \cite{BOUCHER20151}. In simple terms, the approach uses one fold as the testing set and all other folds as the training set. This is repeated $k$ times until all folds have been used as the test set. Finally, the average performance of the model is calculated from the results obtained in each fold, considering the error, precision, or, in the case of regression, the coefficient of determination ($R^2$). In contrast to the Holdout method described previously, it is not necessary to retain a portion of the data for later testing, allowing for a more complete training process in exchange for leaking data to the model.

Cross-validation is less sensitive than the holdout method and is widely used to find and optimize hyperparameters of ML models, resulting in better generalization performance. Typically, a value of $k=10$ is chosen. Although higher values of $k$ may better estimate the real error, it can also increase computational cost, therefore, lower values of $k$ are mostly used for large datasets. \cite{raschka2019python}. In this study, a $k$ of $10$ is used, since our datasets are small and we have moderate computational power.

The leave-one-out method was introduced and formalized by \cite{stone1974cross} as a particular case of cross-validation where the number of folds is equal to the number of samples $n$.
In this method, each sample is used once as a test set, while the remaining set of $n-1$ samples is used for training, and the trained model is used to make a prediction on the held-out sample.
At the end, $n$ evaluation scores (mostly the error) are obtained. The final performance of the model is the average of these $n$ scores.
The objective is to robustly evaluate a model's generalization ability, especially with small datasets \cite{stone1974cross}.
This approach makes maximum use of the data for training. However, it is computationally expensive for large datasets.

A comparison was made using all three methods, and the best train and test subset proportions were chosen based on $R^2$, MSE, and overfitting. The datasets used derive from the previous analysis, where the combination of the SG filter and outlier removal resulted in the best outcomes in all cases. For this analysis, the C contents in Oxisol were evaluated using a PLS regression model, with the $n\_components$ interval ranging from 0 to 40. The results are presented in Table \ref{tab:splitcarbolato}. The in Inceptisol contentswere also analyzed, although a choice was made to keep only one example to maintain consistency in flow and visual clarity.

\subsection{Regression model building}
\label{subsec:regressor_building}

For model training, PLS, SVR, Ridge, RF models, and Stacking Ensemble Learning approaches were utilized. The selection of these algorithms was guided by a comparative analysis among different Pipelines, using cross-validation and grid search, considering metrics such as $R^2$ and the susceptibility to overfitting. The process of creating Stacking Ensemble Learning models first involved training individual models with their respective hyperparameters, which were optimized using a grid search combined with cross-validation with $k=10$ folds. The optimized estimators were then stored in a list. Finally, the Stacking Ensemble Learning scheme was implemented, using the best models as base estimators and employing linear regression as the meta-model to combine the base model predictions.

This study conducted a comparative analysis of various modeling Pipelines to estimate the contents of C and N in Oxisol and Inceptisol samples. This strategy aims to evaluate the influence of different training approaches on the predictive performance of the models, allowing for the identification of the most effective approaches for each soil type and component analyzed. Table \ref{tab:config_pipelines} presents the pipeline settings for building predictive models, detailing the additional configuration steps (if necessary) and the algorithms used.

\begin{table}[!htbp]
\centering
\footnotesize
\caption{Settings of the  Evaluated Model Pipelines.}
\label{tab:config_pipelines}
\renewcommand{\arraystretch}{1.8}
\begin{tabular}{ccc}
\hline
\textbf{Pipeline} & \textbf{Setting} & \textbf{Algorithms} \\ \hline
Pipeline 1 & No additional steps & SVR \\
Pipeline 2 & Standardization & SVR \\ 
Pipeline 3 & \begin{tabular}[c]{@{}c@{}}Standardization  with PCA (99\% variance)\end{tabular} & SVR \\ 
Pipeline 4 & No additional steps & PLS \\ 
Pipeline 5 & No additional steps & RF \\ 
Pipeline 7 & \begin{tabular}[c]{@{}c@{}}Stacking Ensemble: SVR + Ridge\end{tabular} & SVR, Ridge \\ 
Pipeline 8 & \begin{tabular}[c]{@{}c@{}}Stacking Ensemble: PLS + SVR\end{tabular} & PLS, SVR \\ 
Pipeline 9 & \begin{tabular}[c]{@{}c@{}}Stacking Ensemble: PLS + Ridge + SVR\end{tabular} & PLS, Ridge, SVR \\ \hline
\end{tabular}
\end{table}

Pipelines 1 to 3 employed the SVR algorithm with varying levels of additional preprocessing, ranging from the absence of extra steps to the combined application of standardization and dimensionality reduction. Pipeline 4, on the other hand, utilized PLS, allowing for a comparison of the algorithm's performance with prior data standardization. Pipelines 5 to 9 adopted the Stacking Ensemble Learning strategy, combining base models such as PLS, SVR, and Ridge, aiming to enhance robustness and predictive capability. The selection of these parameters was done empirically, in a study not yet published. They were delineated to balance the search for performance and computational feasibility.
Table \ref{quad:optimization_parameters} describes the hyperparameter search space, including the intervals and discrete values tested during the optimization step.

\begin{table}[!htb]
\centering
\caption{Optimal settings for models SVR, PLS, Ridge and Random Forest}
\label{quad:optimization_parameters}
\begin{tabular}{lcc}
\hline
\textbf{Algorithm} & \textbf{Parameter} & \textbf{Interval} \\
\hline
\multirow{4}{*}{SVR} 
& C & 0,01; 0,1; 1; 10; 100 \\
& epsilon & 0,001; 0,01; 0,1; 0,2; 0,3 \\
& gamma & 'scale'; 'auto'; 0,001; 0,01; 0,1; 1 \\
& kernel & 'rbf'; 'linear' \\
\hline
\multirow{4}{*}{PLS}
& n\_components & [1,40]\\
& scale & True; False \\
& max\_iter & 500; 1000 \\
& tol & 1e-6; 1e-7; 1e-8 \\
\hline
\multirow{1}{*}{Ridge}
& alpha & 0,01; 0,1; 1,0; 10 \\
\hline
\multirow{5}{*}{Random Forest}
& n\_estimators & 100; 200; 300 \\
& max\_depth & None; 10; 20; 30 \\
& min\_samples\_split & 2; 5; 10 \\
& min\_samples\_leaf & 1; 2; 4 \\
& max\_features & 'sqrt'; 'log2'; None \\
\hline
\end{tabular}
\end{table}

For the SVR, the following parameters were tested for model tuning: the regularization parameter $\mathbf{C}$ (Cost), epsilon ($\epsilon$), gamma ($\gamma$), and Linear and RBF Kernels. Although the wide range of tested values could represent a significant computational cost—especially considering the complexity of SVR, which scales cubically with the number of samples—this approach was necessary to adequately capture the nuances of the spectral dataset, which is known to be complex and high-dimensional.

The PLS model requires optimization focused on the number of latent components ($\mathbf{n\_components}$), a crucial parameter for capturing the covariance between predictor variables and the response variable. The inclusion of convergence parameters ($\mathbf{max\_iter}$ and $\mathbf{tol}$) ensured the numerical stability of the algorithm, while the $\mathbf{scale}$ option allowed for the evaluation of the impact of standard scaling on the data.

For Ridge regression, the algorithm's configuration simplicity allows for focusing on the optimization of the ($\alpha$) parameter, which is fundamental for controlling the balance between bias and variance, thus preventing overfitting in spectral data, as this type of data often exhibits multicollinearity.

The RF required the optimization of multiple parameters that control the complexity of the trees: $\mathbf{n\_estimators}$ (number of trees), $\mathbf{max\_depth}$ (maximum depth), $\mathbf{min\_samples\_split}$ and $\mathbf{min\_samples\_leaf}$ (pruning controls), and\\ $\mathbf{max\_features}$ (number of features considered for each split).

The tested hyperparameter intervals, although computationally demanding in some cases (particularly for SVR), proved justifiable by the results obtained in our previous study, where all algorithms presented satisfactory performance both in calibration and validation. The use of cross-validation and grid search techniques ensured that the models not only fitted the training data reasonably well but also generalized adequately to unseen data, thereby reducing overfitting.

\subsection{Model evaluation}
\label{subsec:model_evaluation}

The main performance evaluation methods used in this study, their mathematical formulations, and application criteria are addressed below.

\begin{itemize}
    \item Coefficient of determination ($R^2$): refers to the proportion of variability in the data that is explained by the regression model, where values closer to 1 indicate a better model fit. The coefficient is frequently used to analyze the model's adequacy \cite{montgomery2010applied}. $R^2$ is given according to Equation \ref{eq:r2}, in which $y_i$ is the actual value of the $i$-th sample, $\hat{y}_i$ is the predicted value, $\bar{y}$ is the mean of the actual values, and $n$ is the number of samples.
    
  \begin{equation}\label{eq:r2}
 R^2 = 1 - \frac{\sum_{i=1}^{n}(y_i - \hat{y}_i)^2}{\sum_{i=1}^{n}(y_i - \bar{y})^2}
 \end{equation}
    
\item Mean Squared Error (MSE): describes how far the points are from a regression line. This is done by calculating the average of the distances of these points from the regression line and squaring that value, according to Equation \ref{eq:mse} \cite{khan2019performance}. MSE quantifies the average of the squares of the differences between the target values ($y_i$) and the predicted values ($\hat{y}_i$) in a statistical model. The quadratic part of the equation is essential to eliminate negative values and emphasize very large errors.

\begin{equation}\label{eq:mse}
 \text{MSE} = \frac{1}{n}\sum_{i=1}^{n}(y_i - \hat{y}_i)^2
 \end{equation}

\item Root Mean Squared Error (RMSE): is obtained by taking the square root of the MSE, according to Equation \ref{eq:rmse} to yield the RMSE. The smaller the RMSE value, the better the model performance is considered \cite{louw2010robust}.
 
In \textbf{Equation \ref{eq:mse}}, $n$ denotes the number of observations, $y_i$ corresponds to the actual value of the $i$-th observation, and $\hat{y}_i$ represents the value predicted by the model for the same observation. MSE quantifies the average of the squares of the differences between the actual values ($y_i$) and the predicted values ($\hat{y}_i$) in a statistical model. The quadratic part of the equation is essential to eliminate negative values and gives emphasis to very large errors.

\begin{equation}\label{eq:rmse}
\text{RMSE} = \sqrt{MSE}
\end{equation}

\item RMSE and MSE: These two metrics have a monotonic relationship (as one variable increases, the other always varies in the same direction). From the RMSE, it is possible to analyze how the model fits the data in an interpretation with the same unit as the $y$ variable. An $RMSE  = 0$ means the model predicts the data perfectly \cite{khan2019performance}.

\item Mean Absolute Error (MAE): is calculated by averaging the absolute loss values (Equation \ref{eq:mae}).  This metric is useful because it is expressed in the same unit as the $y$ variable and is less sensitive to extreme values when compared to RMSE \cite{willmott2005advantages}. The MAE provides a more direct interpretation of the model's average error, being recommended for predictive performance evaluation \cite{chai2014root}. It is also useful because it is expressed in the same unit as the $y$ variable and is less sensitive to extreme values when compared to RMSE \cite{willmott2005advantages}.
    
\begin{equation}
\label{eq:mae}
\text{MAE} = \frac{1}{n} \sum_{i=1}^{n} |y_i - \hat{y}_i|
\end{equation}

in which $|y_i - \hat{y}_i|$ represents the absolute deviation between the actual value and the predicted value for each observation. The summation is the set of all these deviations, while the division by $n$ calculates the average value. 

 \item Ratio of Performance Deviation (RPD): is a non-dimensional metric widely used to evaluate NIR spectroscopy. This measure is given by the ratio between the SD of the target values and the RMSE, according to Equation \ref{eq:rpd}. RPD is a metric that provides important insights into the model's predictive capability. RPD values between 1.5 and 2.0 are considered insufficient for practical applications, while values between 2.0 and 2.5 indicate approximate quantitative predictions. Models with RPD between 2.5 and 3.0 are classified as good predictors, and values above 3.0 are considered excellent for prediction. Therefore, the higher the RPD value, the better the model's predictive capability \cite{saeys2005potential}.  In short, RPD allows analyzing the RMSE based on the Standard Deviation obtained through Equation \ref{eq:dp} \cite{esbensen2014rpd}.

\begin{equation}
\label{eq:rpd}
\text{RPD} = \frac{\sigma(y)}{\text{RMSE}}
\end{equation}
\end{itemize}

\section{Results and Discussion}
\label{sec:results_and_discussion}

In this section, the evaluation of the Validation methods is presented first. Then, we present the differences resulting from applying multiple preprocessing methods in combination with each other, as well as the absence of preprocessing (raw spectra). To ensure a consistent, computationally efficient, and single-hyperparameter comparison, predictions in this step were generated using only the PLS regression model. PLS was also selected due to inherently including data preparation techniques like standardization and dimensional reduction. The data was split into 70\% training and 30\% testing sets using the Holdout method combined with the Kennard-Stone algorithm for this purpose.

After the comparison of preprocessing methods, the prepared datasets were subjected to a Pipeline comparison analysis, where the final results were obtained.

\subsection{Validation Methods Comparison}

In this section, results recurring from an analysis with PLS regression and multiple data separation techniques are presented. Methods such as Holdout (with Kennard-Stone algorithm), K-Fold and Leave-One-Out are tested and measured by R² and MSE.

\begin{table}[]
\caption{Comparison of Validation Methods by MSE and R² for Oxisol Samples using PLS to Estimate Carbon}
\centering
\begin{tabular}{cccc}
\hline
\textbf{Validation Method}               & \textbf{Test Size / Strategy} & \textbf{MSE}  & \textbf{R²}   \\ \hline
K-Fold                                   & 10 Folds                      & 0.232 ± 0.113 & 0.853 ± 0.039 \\ \hline
Leave-One-Out                            & 1 Sample                      & 0.245         & 0.854         \\ \hline
\multirow{7}{*}{Holdout (Kennard-Stone)} & 40\%                          & 0.294         & 0.796         \\
                                         & 35\%                          & 0.293         & 0.796         \\
                                         & 30\%                          & 0.203         & 0.855         \\
                                         & 25\%                          & 0.176         & 0.882         \\
                                         & 20\%                          & 0.183         & 0.867         \\
                                         & 15\%                          & 0.165         & 0.873         \\
                                         & 10\%                          & 0.159         & 0.820         \\ \hline
\end{tabular}
\label{tab:splitcarbolato}
\end{table}

The results presented in Table \ref{tab:splitcarbolato} demonstrated a clear dependency of the MSE on the test set size. Regression methods are sensitive to the representativeness of the training data. When the test set size is reduced below 20\% (Holdout KS 15\%, 10\%), MSE values of 0.165 and 0.159 are observed, which indicate a drastic improvement in performance. Therefore, these values should be interpreted as artificial or misleading. This behavior suggests that, with few samples in the test set, the Kennard-Stone method selects points that do not adequately challenge the model, creating a false impression of high precision. For test sizes above 30\% (Holdout KS 30\% to 40\%), the MSE stabilizes at values close to those obtained with cross-validation (K-Fold: 0.232; LOO: 0.244), indicating that these configurations better capture the model's true generalization ability in this study. Leave-One-Out (LOO), although computationally expensive, is the most robust way to obtain the MSE. A phenomenon observed in this work's experiments is the abrupt transition ($\approx 31\%$ ) in the MSE between the 35\% split (MSE = 0.293) and the 30\% split (MSE = 0.202). This change suggests that, for this specific dataset, 30\% represents a minimum threshold for the test set to be statistically representative. In conclusion, the analysis suggests that, in this study, the PLS training to estimate C contents for the Oxisol samples can be performed using Holdout with the Kennard-Stone algorithm, employing a 30\% test set and a 70\% training set, or by using the K-Fold strategy. The same analysis is performed to estimate N in Oxisol, as well as C and N in Cambisol.

\subsection{Results corresponding to Oxisol Samples}\label{results_oxisol}

In this Section we present the preprocessing methods and model calibration results corresponding to the Oxisol soils for the prediction of C and N content.

\subsubsection{Carbon Content in Oxisol Soils}

In Table \ref{tab:pre_carbonolato}, referring to the preprocessing results to the prediction of the C content in Oxisol, it is shown that the combination SG + Outlier achieved the best results. The validation ($R^2=0.86$; $RMSE=0.45$; $RPD=2.63$; $MAE=0.33$) and calibration ($R^2=0.91$; $RMSE=0.40$; $RPD=3.31$; $MAE= 0.31$) values indicate a robust model, both with $RPD > 2.50$.

\begin{table}[]
\caption{Comparison between preprocessing operations in Oxisol samples aiming at Carbon content}
\centering
\label{tab:pre_carbonolato}
\begin{tabular}{cccccc}
\hline
\textbf{Dataset} &
  \textbf{Preprocessing} &
  \textbf{R²} &
  \textbf{RMSE} &
  \textbf{RPD} &
  \textbf{MAE} \\ \hline
 &
  RAW &
  0.70 &
  0.75 &
  1.82 &
  0.57 \\
 &
  TRIM &
  0.71 &
  0.73 &
  1.87 &
  0.54 \\
 &
  SG &
  0.78 &
  0.64 &
  2.13 &
  0.48 \\
 &
  SNV &
  0.41 &
  1.04 &
  1.31 &
  0.78 \\
 &
  MSC &
  0.22 &
  1.20 &
  1.13 &
  0.84 \\
 &
  LOG &
  0.60 &
  0.86 &
  1.58 &
  0.64 \\
 &
  LOG → SG &
  0.82 &
  0.58 &
  2.33 &
  0.47 \\
 &
  SG → LOG → MC &
  0.67 &
  0.76 &
  2.04 &
  0.50 \\
 &
  TRIM → SG &
  0.78 &
  0.64 &
  2.13 &
  0.48 \\
\multirow{-10}{*}{Validation} &
  \cellcolor[HTML]{FFFFFF}SG → Outlier &
  \cellcolor[HTML]{C0C0C0}0.86 &
  \cellcolor[HTML]{C0C0C0}0.45 &
  \cellcolor[HTML]{C0C0C0}2.63 &
  \cellcolor[HTML]{C0C0C0}0.33 \\ \hline
 &
  RAW &
  0.84 &
  0.75 &
  2.00 &
  0.61 \\
 &
  TRIM &
  0.85 &
  0.60 &
  2.58 &
  0.49 \\
 &
  SG &
  0.75 &
  0.77 &
  2.08 &
  0.61 \\
 &
  SNV &
  0.93 &
  0.64 &
  1.66 &
  0.71 \\
 &
  MSC &
  0.64 &
  0.83 &
  2.40 &
  0.52 \\
 &
  LOG &
  0.79 &
  0.74 &
  1.96 &
  0.64 \\
 &
  LOG → SG &
  0.79 &
  0.74 &
  1.96 &
  0.64 \\
 &
  SG → LOG → MC &
  0.64 &
  0.83 &
  2.40 &
  0.52 \\
 &
  TRIM → SG &
  0.84 &
  0.63 &
  2.48 &
  0.52 \\
\multirow{-10}{*}{Calibração} &
  \cellcolor[HTML]{FFFFFF}SG → Outlier &
  \cellcolor[HTML]{C0C0C0}0.91 &
  \cellcolor[HTML]{C0C0C0}0.40 &
  \cellcolor[HTML]{C0C0C0}3.31 &
  \cellcolor[HTML]{C0C0C0}0.31 \\ \hline
\end{tabular}
\end{table}

Apart from SG smoothing combined with outlier removal techniques, isolated SG and LOG $\to$ SG also showed competitive results. In contrast, SNV and MSC were clearly inadequate for the content of C in this soil, resulting in validation models with inferior predictive power ($R^2 < 0.45$) and RPD close to 1.2.

Table \ref{tab:resultados_latossolo_carbono} presents the results of model calibration using different Pipelines, as described in Section \ref{sec:maerial_and_methods}, for predicting C in Oxisol soils.

\begin{table}[!htbp]
\centering
\caption{Comparative results of models for Carbon in Oxisol}
\label{tab:resultados_latossolo_carbono}
\renewcommand{\arraystretch}{1.4}
\setlength{\tabcolsep}{4pt}
\begin{tabular}{lcccccccc}
\hline
\multirow{2}{*}{} &
\multicolumn{4}{c}{\textbf{Calibration}} &
\multicolumn{4}{c}{\textbf{Validation}} \\ \hline
& R² & RMSE & MAE & RPD & R² & RMSE & MAE & RPD \\ \hline
\multicolumn{9}{l}{\textbf{SVR}} \\ 
Pipeline 1 & 0.85 & 0.50 & 0.32 & 2.60 & 0.82 & 0.52 & 0.36 & 2.35 \\ 
\rowcolor[HTML]{CCCCCC}
Pipeline 2 & 0.90 & 0.42 & 0.29 & 3.11 & 0.91 & 0.36 & 0.31 & 3.39 \\ 
Pipeline 3 & 0.93 & 0.35 & 0.18 & 3.76 & 0.87 & 0.44 & 0.33 & 2.77 \\ \hline
\multicolumn{9}{l}{\textbf{PLS}} \\ 
Pipeline 4 & 0.91 & 0.40 & 0.31 & 3.31 & 0.86 & 0.45 & 0.33 & 2.64 \\ \hline
\multicolumn{9}{l}{\textbf{\begin{tabular}[c]{@{}l@{}} Random Forest\end{tabular}}} \\ 
Pipeline 5 & 0.96 & 0.27 & 0.17 & 4.87 & 0.85 & 0.45 & 0.33 & 2.60 \\ \hline
\multicolumn{9}{l}{\textbf{\begin{tabular}[c]{@{}l@{}}Stacking Ensembles\end{tabular}}} \\ 
Pipeline 6 & 0.91 & 0.39 & 0.30 & 3.38 & 0.87 & 0.43 & 0.33 & 2.75 \\ 
Pipeline 7 & 0.92 & 0.38 & 0.28 & 3.48 & 0.88 & 0.40 & 0.32 & 2.95 \\ 
Pipeline 8 & 0.91 & 0.40 & 0.30 & 3.32 & 0.88 & 0.40 & 0.31 & 2.94 \\ 
Pipeline 9 & 0.91 & 0.40 & 0.30 & 3.32 & 0.88 & 0.40 & 0.31 & 2.94 \\ \hline
\end{tabular}
\end{table}

The analysis of the results presented in Table \ref{tab:resultados_latossolo_carbono} reveals differences in Calibration and Validation performance among the models tested for the prediction of C in Oxisol. In the Calibration dataset, Pipeline 5 (RF) achieved the highest $R^2 = 0.96$ and the lowest errors, with $RMSE = 0.27$ and $MAE = 0.17$. This resulted in the highest Calibration RPD ($4.87$). However, in Validation, the $R^2$ of Pipeline 5 drops to $0.85$ and the RMSE increases to $0.45$, with the RPD falling to $2.60$. This divergence between Calibration and Validation indicates that the model adapted very specifically to the training data and can be considered a moderate case of overfitting. In contrast, Pipeline 2 (SVR) demonstrates a greater capacity for generalization, achieving the highest Validation $R^2$ of $0.91$. This performance correlates with the lowest prediction errors in the test set ($RMSE = 0.36$ and $MAE = 0.31$), also resulting in the highest Validation RPD of $3.39$. The small difference between the Calibration $R^2$ ($0.90$) and Validation $R^2$ ($0.91$) for Pipeline 2 indicates that the model achieved consistent predictive capability across the stages and, consequently, very low overfitting. The RPD of $3.39$ signifies model robustness. Pipeline 4 (PLS), although the simplest, presents an $R^2$ of $0.91$ in Calibration and $0.86$ in Validation, combined with an RMSE of $0.45$ in Validation. Therefore, it can be considered an intermediate model in this study. The Stacking Ensemble Pipelines exhibit consistent performance.

\subsubsection{Nitrogen Content in Oxisol Soils}

In Table  \ref{tab:pre_nitrogeniolato},to the preprocessing results to the prediction of N in the Oxisol soil samples, and again (such as in C content) the SG + Outlier combination achieved the best results. Similar results were observed (Validation: $R^2=0.84$, $RMSE=0.04$, $RPD=2.51$, $MAE=0.03$; Calibration: $R^2=0.88$, $RPD=2.91$, $MAE=0.04$).

\begin{table}[!htbp]
\caption{Comparison between preprocessing operations in Oxisol samples aiming at Nitrogen content}
\label{tab:pre_nitrogeniolato}
\centering

\begin{tabular}{cccccc}
\hline
\textbf{Dataset} & \textbf{Preprocessing} & \textbf{R²} & \textbf{RMSE} & \textbf{RPD} & \textbf{MAE} \\ \hline

\multirow{10}{*}{\centering Validation}
 & RAW SPECTRUM & 0.65 & 0.07 & 1.69 & 0.05 \\ 
 & TRIM & 0.67 & 0.07 & 1.75 & 0.05 \\ 
 & SG & 0.75 & 0.06 & 2.00 & 0.05 \\ 
 & SNV & 0.27 & 0.10 & 1.17 & 0.07 \\ 
 & MSC & 0.15 & 0.11 & 1.08 & 0.08 \\ 
 & LOG & 0.66 & 0.07 & 1.72 & 0.06 \\ 
 & LOG $\to$ SG & 0.71 & 0.07 & 1.87 & 0.06 \\ 
 & SG $\to$ LOG $\to$ MC & 0.81 & 0.06 & 2.30 & 0.05 \\ 
 & TRIM $\to$ SG & 0.73 & 0.06 & 1.94 & 0.05 \\ 
 & SG $\to$ Outlier & \cellcolor[HTML]{C0C0C0}0.84 & \cellcolor[HTML]{C0C0C0}0.04 & \cellcolor[HTML]{C0C0C0}2.52 & \cellcolor[HTML]{C0C0C0}0.03 \\ \hline

\multirow{10}{*}{\centering Calibration}
 & RAW SPECTRUM & 0.74 & 0.07 & 1.97 & 0.06 \\ 
 & TRIM & 0.83 & 0.06 & 2.41 & 0.05 \\ 
 & SG & 0.68 & 0.08 & 1.77 & 0.06 \\ 
 & SNV & 0.76 & 0.07 & 2.02 & 0.06 \\ 
 & MSC & 0.71 & 0.07 & 1.87 & 0.06 \\ 
 & LOG & 0.81 & 0.06 & 2.30 & 0.05 \\ 
 & LOG $\to$ SG & 0.71 & 0.07 & 1.87 & 0.06 \\ 
 & SG $\to$ LOG $\to$ MC & 0.81 & 0.06 & 2.30 & 0.05 \\ 
 & TRIM $\to$ SG & 0.80 & 0.06 & 2.26 & 0.05 \\ 
 & SG $\to$ Outlier & \cellcolor[HTML]{C0C0C0}0.88 & \cellcolor[HTML]{C0C0C0}0.04 & \cellcolor[HTML]{C0C0C0}2.91 & \cellcolor[HTML]{C0C0C0}0.04 \\ \hline

\end{tabular}
\end{table}

Techniques like isolated SG and LOG $\to$ SG also showed competitive results for N. SNV and MSC were again clearly inadequate for the soil, resulting in validation models with inferior adequacy ($R^2 < 0.30$) and RPD values close to 1.0.

Table \ref{tab:resultados_latossolo_nitrogenio} presents the results of model calibration using different Pipelines, as described in Section \ref{sec:maerial_and_methods}, for predicting C in Inceptisol soils.

\begin{table}[!htbp]
\centering
\caption{Comparative results of models for Nitrogen in Oxisol}
\label{tab:resultados_latossolo_nitrogenio}
\renewcommand{\arraystretch}{1.4}
\setlength{\tabcolsep}{4pt}
\begin{tabular}{lcccccccc}
\hline
\multirow{2}{*}{} &
\multicolumn{4}{c}{\textbf{Calibration}} &
\multicolumn{4}{c}{\textbf{Validation}} \\ \hline
& R² & RMSE & MAE & RPD & R² & RMSE & MAE & RPD \\ \hline
\multicolumn{9}{l}{\textbf{SVR}} \\ 
Pipeline 1 & 0.93 & 0.03 & 0.02 & 3.92 & 0.52 & 0.07 & 0.04 & 1.44 \\ 
\rowcolor[HTML]{CCCCCC}
Pipeline 2 & 0.95 & 0.03 & 0.01 & 4.38 & 0.89 & 0.03 & 0.03 & 3.01 \\ 
Pipeline 3 & 0.93 & 0.03 & 0.01 & 3.67 & 0.85 & 0.04 & 0.03 & 2.54 \\ \hline
\multicolumn{9}{l}{\textbf{PLS}} \\ 
Pipeline 4 & 0.91 & 0.04 & 0.03 & 3.32 & 0.84 & 0.04 & 0.03 & 2.49 \\ \hline
\multicolumn{9}{l}{\textbf{\begin{tabular}[c]{@{}l@{}} Random Forest\end{tabular}}} \\ 
Pipeline 5 & 0.96 & 0.02 & 0.02 & 4.97 & 0.81 & 0.04 & 0.03 & 2.31 \\ \hline
\multicolumn{9}{l}{\textbf{\begin{tabular}[c]{@{}l@{}}Stacking Ensembles\end{tabular}}} \\ 
Pipeline 6 & 0.88 & 0.04 & 0.03 & 2.86 & 0.85 & 0.04 & 0.03 & 2.55 \\ 
Pipeline 7 & 0.94 & 0.03 & 0.02 & 4.04 & 0.88 & 0.03 & 0.03 & 2.87 \\ 
Pipeline 8 & 0.94 & 0.03 & 0.02 & 4.05 & 0.88 & 0.04 & 0.03 & 2.83 \\ 
Pipeline 9 & 0.94 & 0.03 & 0.02 & 4.04 & 0.88 & 0.04 & 0.03 & 2.84 \\ \hline
\end{tabular}
\end{table}

The analysis of the Pipelines for N in Oxisol, based on the metrics from Table \ref{tab:resultados_latossolo_nitrogenio} (Calibration and Validation $R^2$, RMSE, MAE, and RPD), reveals that Pipeline 1 shows the largest performance discrepancy: it has a high Calibration $R^2$ of $0.93$ and RPD of $3.92$, but yields the worst Validation $R^2$ ($0.52$) and RPD ($1.44$), with the highest RMSE ($0.07$). Pipeline 2 outperforms the other SVR models in Validation, reaching an $R^2$ of $0.89$ and an RPD of $3.01$. It also boasts the highest Calibration $R^2$ ($0.95$) and RPD ($4.38$) within this group, and its Validation RMSE ($0.03$) is the lowest. Pipeline 3 presents intermediate results, with a Validation $R^2$ of $0.85$ and RPD of $2.54$.Pipeline 4 (PLS) achieves a Calibration $R^2$ of $0.91$ and RPD of $3.32$. In Validation, its $R^2$ is $0.84$ and RPD of $2.49$. This pipeline exhibits consistency, with stable RMSE ($0.04$) and MAE ($0.03$) between Calibration and Validation, being one of the models with the smallest drop in $R^2$ between the stages. Pipeline 5 (RF) shows the best fit in Calibration among all models, with an $R^2$ of $0.96$ and the maximum RPD ($4.97$), and the lowest RMSE ($0.02$). However, its Validation performance drops sharply to an $R^2$ of $0.81$ and RPD of $2.31$, indicating significant overfitting. Its Validation RMSE and MAE ($0.04$ and $0.03$) are higher than those obtained by the SVR and Stacking models. Pipeline 6 shows the lowest performance in the Stacking Ensembles group, with a Calibration $R^2$ of $0.88$ and RPD of $2.86$, but it demonstrates the highest consistency among all configurations, maintaining a Validation $R^2$ of $0.85$ and RPD of $2.55$. Pipelines 7, 8, and 9 present nearly identical Calibration results ($R^2 \sim 0.94$ and RPD $\sim 4.04$) and also very close Validation results ($R^2 \sim 0.88$ and RPD $\sim 2.8$).Therefore, Pipeline 2 again registers the strongest validation performance, this time for N, with the highest $R^2$ ($0.89$) and RPD ($3.01$) and a low RMSE ($0.03$). While Pipeline 5 achieves the best Calibration ($R^2: 0.96$, RPD: $4.97$), Pipeline 6 stands out for its greater stability and minimal variation in metrics between Calibration and Validation, indicating the least overfitting.

\subsection{Results corresponding to Inceptisol Samples}

In this Section we present the preprocessing methods and model calibration results corresponding to the Oxisol soils for the prediction of C and N content.

\subsubsection{Carbon Content in Inceptisol Soils}

In Table \ref{tab:pre_cambissolo_carbono} corresponding to the preprocessing results to the prediction of C content in Inceptisol soil samples. The SG + Outlier combination achieved the best results (Validation: $R^2=0.75$, $RMSE=0.61$, $RPD=2.00$, $MAE=0.53$; Calibration: $R^2=0.83$, $RMSE=0.53$, $MAE=0.43$,$RPD=2.42$). 

\begin{table}[!h]
\centering
\caption{Comparison between preprocessing operations in Inceptisol  samples aiming at Carbon content.}
\label{tab:pre_cambissolo_carbono}
\begin{tabular}{cccccc}
\hline
\textbf{Dataset} & \textbf{Preprocessing} & \textbf{R²} & \textbf{RMSE} & \textbf{MAE} & \textbf{RPD} \\ \hline

\multirow{10}{*}{\centering Validation}
 & RAW SPECTRUM & 0.55 & 0.93 & 0.68 & 1.48 \\ 
 & TRIM & 0.57 & 0.91 & 0.70 & 1.52 \\ 
 & SNV & 0.58 & 0.90 & 0.66 & 1.54 \\ 
 & MSC & 0.50 & 0.97 & 0.70 & 1.42 \\ 
 & LOG & 0.36 & 1.10 & 0.85 & 1.25 \\ 
 & SG & 0.60 & 0.88 & 0.69 & 1.57 \\ 
 & LOG $\to$ SG & 0.51 & 0.97 & 0.74 & 1.43 \\ 
 & SG $\to$ LOG $\to$ MC & 0.59 & 0.88 & 0.69 & 1.56 \\ 
 & TRIM $\to$ SG & 0.59 & 0.89 & 0.71 & 1.55 \\ 
 & SG $\to$ Outlier & \cellcolor[HTML]{C0C0C0}0.75 & \cellcolor[HTML]{C0C0C0}0.61 & \cellcolor[HTML]{C0C0C0}0.53 & \cellcolor[HTML]{C0C0C0}2.00 \\ \hline

\multirow{10}{*}{\centering Calibration}
 & RAW SPECTRUM & 0.74 & 0.71 & 0.53 & 1.97 \\ 
 & TRIM & 0.78 & 0.66 & 0.50 & 2.12 \\ 
 & SNV & 0.73 & 0.72 & 0.53 & 1.92 \\ 
 & MSC & 0.71 & 0.75 & 0.54 & 1.84 \\ 
 & LOG & 0.64 & 0.83 & 0.61 & 1.67 \\ 
 & SG & 0.79 & 0.63 & 0.46 & 2.20 \\ 
 & LOG $\to$ SG & 0.73 & 0.72 & 0.55 & 1.93 \\ 
 & SG $\to$ LOG $\to$ MC & 0.75 & 0.70 & 0.53 & 1.99 \\ 
 & TRIM $\to$ SG & 0.79 & 0.63 & 0.47 & 2.19 \\ 
 & SG $\to$ Outlier & \cellcolor[HTML]{C0C0C0}0.83 & \cellcolor[HTML]{C0C0C0}0.53 & \cellcolor[HTML]{C0C0C0}0.43 & \cellcolor[HTML]{C0C0C0}2.42 \\ \hline

\end{tabular}
\end{table}

As shown in Table \ref{tab:pre_cambissolo_carbono} the models exhibited inferior metrics compared to those of Oxisol. Even so, the SG + Outlier technique remained the best option, although The validation RPD ($2.002$) was significantly below the $2.5$ threshold.  Similar to the Oxisol samples, the techniques combining SG smoothing with outlier removal (SG + Outlier) also demonstrated the best performance in Inceptisol soils compared to other preprocessing methods in this study. Interestingly, the raw spectra showed competitive performance in this soil type, outperforming several of the preprocessing techniques used.

\begin{table}[!htbp]
\centering
\caption{Comparative results of models for Carbon in Inceptisol}
\label{tab:resultados_cambissolo_carbono}
\renewcommand{\arraystretch}{1.4}
\setlength{\tabcolsep}{4pt}
\begin{tabular}{lcccccccc}
\hline
\multirow{2}{*}{} &
\multicolumn{4}{c}{\textbf{Calibration}} &
\multicolumn{4}{c}{\textbf{Validation}} \\\hline
& R² & RMSE & MAE & RPD & R² & RMSE & MAE & RPD \\ \hline
\multicolumn{9}{l}{\textbf{SVR}} \\ 
Pipeline 1 & 0.73 & 0.66 & 0.43 & 1.92 & 0.61 & 0.76 & 0.52 & 1.61 \\ 
Pipeline 2 & 0.84 & 0.51 & 0.41 & 2.51 & 0.76 & 0.60 & 0.51 & 2.05 \\ 
Pipeline 3 & 0.69 & 0.71 & 0.51 & 1.80 & 0.52 & 0.85 & 0.64 & 1.44 \\ \hline
\multicolumn{9}{l}{\textbf{PLS}} \\ 
Pipeline 4 & 0.86 & 0.48 & 0.37 & 2.66 & 0.77 & 0.59 & 0.50 & 2.07 \\ \hline
\multicolumn{9}{l}{\textbf{\begin{tabular}[c]{@{}l@{}}Random Forest\end{tabular}}} \\ 
Pipeline 5 & 0.95 & 0.28 & 0.20 & 4.60 & 0.59 & 0.78 & 0.62 & 1.57 \\ \hline
\multicolumn{9}{l}{\textbf{\begin{tabular}[c]{@{}l@{}}Stacking Ensembles\end{tabular}}} \\ 
\rowcolor[HTML]{CCCCCC}
Pipeline 6 & 0.85 & 0.49 & 0.39 & 2.59 & 0.79 & 0.57 & 0.48 & 2.16 \\ 
Pipeline 7 & 0.84 & 0.51 & 0.41 & 2.51 & 0.76 & 0.60 & 0.51 & 2.04 \\ 
Pipeline 8 & 0.85 & 0.50 & 0.39 & 2.57 & 0.77 & 0.59 & 0.50 & 2.09 \\ 
Pipeline 9 & 0.85 & 0.50 & 0.39 & 2.57 & 0.77 & 0.59 & 0.50 & 2.08 \\ \hline
\end{tabular}
\end{table}

The analysis of the results for C in Inceptisol  reveals that overall, the models showed inferior metrics compared to those obtained for C and the Oxisols. Pipeline 1 registered the weakest validation performance, with the lowest $R^2$ ($0.34$) and an RPD of $1.34$, indicating a model with low predictive power. Pipeline 3 was marginally better, with $R^2=0.45$ and an RPD of $1.44$. In contrast, Pipeline 2 achieved the best validation performance of the SVR group with $R^2$ of $0.60$ and an RPD of $1.61$.Pipeline 4 (PLS) demonstrated notable consistency, maintaining very close calibration ($R^2=0.79$, RPD $2.18$) and validation metrics ($R^2=0.76$, RPD $2.04$), making it one of the models with the smallest performance variation between stages. Its validation RPD ($2.04$) is the second-best across all models. Pipeline 5 (RF) once again achieved the best fit in calibration ($R^2=0.94$, RPD $4.24$), but suffered a drastic drop in validation ($R^2=0.58$, RPD $1.58$), indicating a high degree of overfitting.The stacking ensemble pipelines showed significantly stronger validation results. Pipeline 6 achieved the best overall validation performance, with the highest $R^2$ ($0.77$) and maximum RPD ($2.07$) among all models, in addition to a low RMSE ($0.05$). Pipelines 7, 8, and 9 presented very similar results to each other ($R^2 \sim 0.73$ and RPD $\sim 1.9$).In conclusion, pipeline 6 was the strongest for predicting N in Inceptisol, with the best validation RPD ($2.07$), although this value is still below the $2.5$ threshold. Pipeline 4 (PLS) and Pipeline 2 (SVR) proved to be robust alternatives with good consistency.

\subsection{Nitrogen Content in Inceptisol Soils}

In Table  \ref{tab:pre_cambissolo_carbono} referring to the preprocessing results to the prediction of N in the Inceptisol soil samples. The SG + Outlier combination achieved the best results (Validation: $R^2=0.64$, $RMSE=0.05$, $RPD=1.660$, $MAE=0.04$; Calibration: $R^2=0.84$, $RMSE=0.04$ $RPD=2.52$, $MAE=0.03$). 

\begin{table}[!h]
\centering
\caption{Comparison between preprocessing operations in Inceptisol samples aiming at Nitrogen content.}
\label{tab:pre_cambissolo_nitrogenio}
\begin{tabular}{cccccc}
\hline
\textbf{Dataset} & \textbf{Preprocessing} & \textbf{R²} & \textbf{RMSE} & \textbf{MAE} & \textbf{RPD} \\ \hline

\multirow{10}{*}{\centering Validation}
 & RAW SPECTRUM & 0.50 & 0.08 & 0.06 & 1.46 \\ 
 & TRIM & 0.53 & 0.07 & 0.05 & 1.46 \\ 
 & SNV & 0.53 & 0.07 & 0.06 & 1.46 \\ 
 & MSC & 0.54 & 0.07 & 0.05 & 1.48 \\ 
 & LOG & 0.13 & 0.10 & 0.07 & 1.07 \\ 
 & SG & 0.63 & 0.07 & 0.05 & 1.64 \\ 
 & LOG $\to$ SG & 0.53 & 0.07 & 0.06 & 1.46 \\ 
 & SG $\to$ LOG $\to$ MC & 0.59 & 0.07 & 0.06 & 1.56 \\ 
 & TRIM $\to$ SG & 0.63 & 0.06 & 0.05 & 1.65 \\ 
 & SG $\to$ Outlier & \cellcolor[HTML]{C0C0C0}0.64 & \cellcolor[HTML]{C0C0C0}0.05 & \cellcolor[HTML]{C0C0C0}0.04 & \cellcolor[HTML]{C0C0C0}1.66 \\ \hline

\multirow{10}{*}{\centering Calibration}
 & RAW SPECTRUM & 0.69 & 0.06 & 0.04 & 1.78 \\ 
 & TRIM & 0.72 & 0.05 & 0.04 & 1.91 \\ 
 & SNV & 0.67 & 0.06 & 0.04 & 1.74 \\ 
 & MSC & 0.66 & 0.06 & 0.04 & 1.73 \\ 
 & LOG & 0.56 & 0.07 & 0.05 & 1.50 \\ 
 & SG & 0.74 & 0.05 & 0.04 & 1.96 \\ 
 & LOG $\to$ SG & 0.67 & 0.06 & 0.04 & 1.74 \\ 
 & SG $\to$ LOG $\to$ MC & 0.74 & 0.05 & 0.04 & 1.97 \\ 
 & TRIM $\to$ SG & 0.75 & 0.05 & 0.04 & 2.00 \\ 
 & SG $\to$ Outlier & \cellcolor[HTML]{C0C0C0}0.84 & \cellcolor[HTML]{C0C0C0}0.04 & \cellcolor[HTML]{C0C0C0}0.03 & \cellcolor[HTML]{C0C0C0}2.52 \\ \hline

\end{tabular}
\end{table}

As shown in Table \ref{tab:pre_cambissolo_nitrogenio} the models again exhibited inferior metrics compared to those of Oxisol. And again, the SG + Outlier technique remained the best option, achieving the highest $R^2$ values N (Validation: $0.635$; Calibration: $0.842$). The validation RPD ($1.656$) indicates that the model is not as accurate and robust as the one developed for Oxisol soils.

\begin{table}[!htbp]
\centering
\caption{Comparative results of models for Nitrogen in Inceptisol}
\label{tab:resultados_cambissolo_nitrogenio}
\renewcommand{\arraystretch}{1.4}
\setlength{\tabcolsep}{4pt}
\begin{tabular}{lcccccccc}
\hline
\multirow{2}{*}{} &
\multicolumn{4}{c}{\textbf{Calibration}} &
\multicolumn{4}{c}{\textbf{Validation}} \\ \hline 
& R² & RMSE & MAE & RPD & R² & RMSE & MAE & RPD \\ \hline
\multicolumn{9}{l}{\textbf{SVR}} \\ 
Pipeline 1 & 0.79 & 0.04 & 0.03 & 2.16 & 0.43 & 0.07 & 0.05 & 1.33 \\ 
Pipeline 2 & 0.79 & 0.04 & 0.03 & 2.18 & 0.75 & 0.05 & 0.04 & 1.98 \\ 
Pipeline 3 & 0.61 & 0.06 & 0.04 & 1.60 & 0.58 & 0.06 & 0.04 & 1.55 \\ \hline
\multicolumn{9}{l}{\textbf{PLS}} \\ 
\rowcolor[HTML]{CCCCCC}
Pipeline 4 & 0.80 & 0.04 & 0.03 & 2.26 & 0.77 & 0.05 & 0.04 & 2.11 \\ \hline
\multicolumn{9}{l}{\textbf{\begin{tabular}[c]{@{}l@{}}Random Forest\end{tabular}}} \\ 
Pipeline 5 & 0.92 & 0.03 & 0.02 & 3.60 & 0.69 & 0.06 & 0.05 & 1.78 \\ \hline
\multicolumn{9}{l}{\textbf{\begin{tabular}[c]{@{}l@{}}Stacking Ensembles\end{tabular}}} \\ 
Pipeline 6 & 0.79 & 0.04 & 0.03 & 2.16 & 0.76 & 0.05 & 0.04 & 2.03 \\ 
Pipeline 7 & 0.80 & 0.04 & 0.03 & 2.22 & 0.77 & 0.05 & 0.04 & 2.09 \\ 
Pipeline 8 & 0.79 & 0.04 & 0.03 & 2.16 & 0.76 & 0.05 & 0.04 & 2.04 \\ 
Pipeline 9 & 0.78 & 0.04 & 0.03 & 2.15 & 0.76 & 0.05 & 0.04 & 2.03 \\ \hline
\end{tabular}
\end{table}

The analysis of the pipeline results for N in Inceptisol, based on the calibration and validation metrics ($R^2$, RMSE, MAE, and RPD) from Table \ref{tab:resultados_cambissolo_nitrogenio}, demonstrates that Pipeline 1 registered the weakest validation performance, showing the lowest validation $R^2$ ($0.43$) and the lowest RPD ($1.33$), with the highest RMSE ($0.07$). Pipeline 2 drastically improved validation performance, achieving an $R^2$ of $0.75$ and an RPD of $1.98$. Pipeline 3 showed intermediate results among the SVR models, with a Validation $R^2$ of $0.58$ and RPD of $1.55$. Pipeline 4 (PLS) achieved the highest calibration $R^2$ and calibration RPD ($2.26$) among SVR and PLS pipelines. Crucially, it attained a validation $R^2$ of $0.77$ and the highest Validation RPD ($2.11$) across all pipelines, along with an RMSE of $0.05$. This model exhibited low overfitting, showing little variation in metrics between calibration and validation. Pipeline 5 (RF) achieved the highest calibration $R^2$ ($0.92$) and RPD ($3.60$), but its validation performance dropped significantly to an $R^2$ of $0.69$ and RPD of $1.78$, indicating the largest performance drop and highest overfitting. Within the stacking ensembles, pipeline 7 showed the best validation performance with an $R^2$ of $0.77$ and RPD of $2.09$. Pipelines 6, 8, and 9 showed very similar validation results ($R^2 \sim 0.76$ and RPD $\sim 2.03$). Therefore, pipeline 4 (PLS) and pipeline 7 (Stacking) recorded the best validation performances, with Pipeline 4 slightly ahead in the RPD ($2.11$ vs $2.09$), while pipeline 5 presented the best fit in calibration but the highest overfitting.

\subsection{Regression analysis} \label{subsubsec:regression}

The analysis of the descriptive statistics for the samples (Tables \ref{tab:cambistats} and \ref{tab:latostats} previously shown in section \ref{subsec:EDA}) reveals key relationships with the predictive model performance. The Oxisol exhibited higher mean values for C ($2.74\%$) and N ($0.21\%$) compared to Inceptisol  ($2.35\%$ and $0.17\%$, respectively), in addition to greater variability (Standard Deviation (SD) of $1.50$ vs $1.39$ for C; $0.13$ vs $0.10$ for N). These characteristics may partially explain the better results obtained in the models for Oxisol. The greater natural variability of the contents in the Oxisol provided a more representative dataset for model training, which is reflected in the higher $R^2$ and RPD values. This is particularly true for N, where the performance difference was more pronounced (RPD of $3.01$ in Oxisol vs $2.11$ in Inceptisol), as the wider range of variation and higher SD in the Oxisol contributed to a superior model. Furthermore, the RMSE and MAE values must be interpreted relative to the component means, offering insight into relative precision. In Inceptisol, the C RMSE ($0.57$) represents approximately $24\%$ of the mean ($2.35\%$), while in Oxisol, the RMSE ($0.36$) corresponds to about $13\%$ of the mean ($2.74\%$), indicating better relative precision. For N, this difference is even more evident, with the RMSE representing $29\%$ of the mean in Inceptisol versus only $17\%$ in Oxisol.

\subsection{Feature Analysis}

To identify the most relevant spectral regions, a variable importance analysis was performed based on the coefficients of the ridge regression model. Larger magnitudes of these coefficients indicate a greater contribution to the prediction: positive coefficients suggest a direct correlation with the response variable, while negative coefficients suggest an inverse relationship. The top 20 spectral bands with the largest coefficient magnitudes (both positive and negative) were selected, allowing for the identification of the most influential wavelengths in predicting C and N contents in Oxisol using Pipeline 2. These models were chosen for this analysis, as opposed to the Inceptisol models, due to their superior performance in this study.

The graphical visualization of the relative feature importance provides insight into the spectral characteristics associated with soil organic C and N. This allows for a better understanding of the relationship between the spectral signal and the measured chemical property.

Figures \ref{fig:top20carbon}, \ref{fig:bot20carbon}, \ref{fig:top20nitrogen}, and \ref{fig:bot20nitrogen} illustrate, respectively, the 20 features of greatest and least importance from the Ridge model for predicting C and N contents in Oxisol. The graphs should be read from left to right. The vertical axis (y-axis) presents the Ridge Regression coefficient values, which represent the relative importance of each feature. The horizontal axis (x-axis) shows the corresponding wavelength bands associated with each feature.

The ridge regression methodology was particularly suitable for this analysis because it does not perform variable selection through complete exclusion (as is the case with Lasso), but rather assigns continuous weights to all features, allowing for a gradual assessment of importance even for less significant variables.

\begin{figure}[!h]
\includegraphics[width=1.0\textwidth]{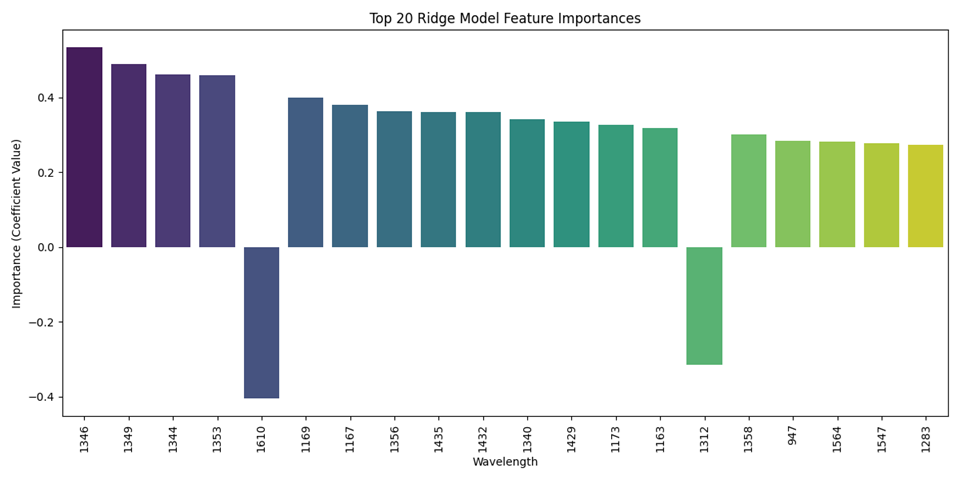}
\caption{20 Largest Coefficients in Absolute Value for Carbon prediction in Oxisol as a function of the respective wavelength}
\label{fig:top20carbon}
\end{figure}

\begin{figure}[!h]
\includegraphics[width=1.0\textwidth]{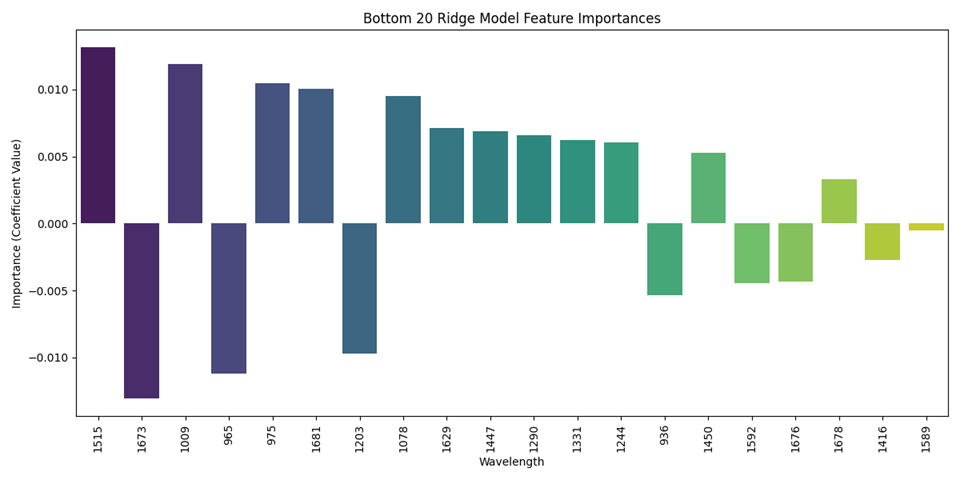}
\caption{20 Smallest Coefficients in Absolute Value for C prediction in Oxisol as a function of the respective wavelength}
\label{fig:bot20carbon}
\end{figure}

\begin{figure}[!h]
\includegraphics[width=1.0\textwidth]{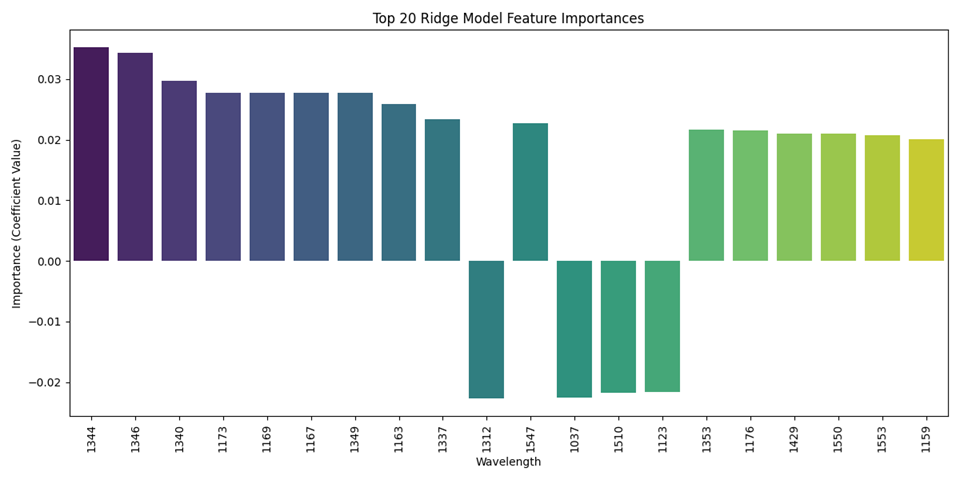}
\caption{20 Largest Coefficients in Absolute Value for Nitrogen prediction in Oxisol as a function of the respective wavelength}
\label{fig:top20nitrogen}
\end{figure}

\begin{figure}[!h]
\includegraphics[width=1.0\textwidth]{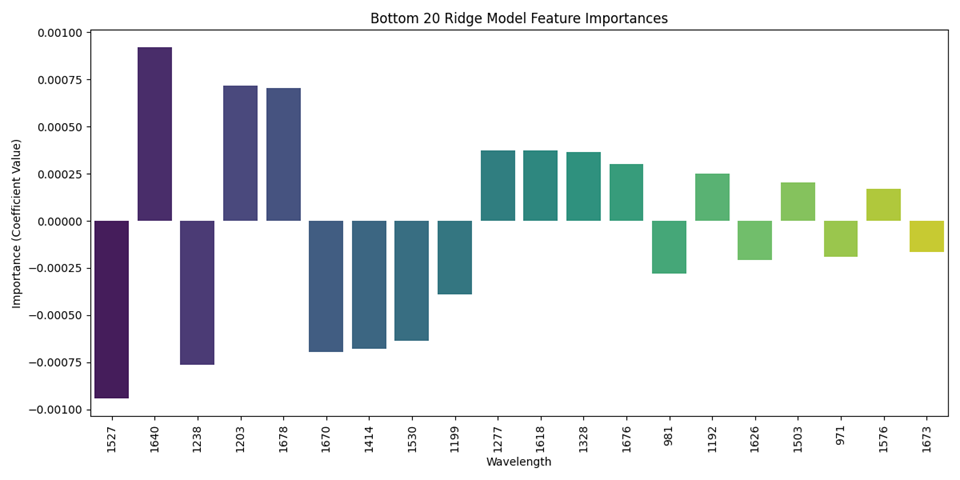}
\caption{20 Smallest Coefficients in Absolute Value for Nitrogen prediction in Oxisol as a function of the respective wavelength}
\label{fig:bot20nitrogen}
\end{figure}

The analysis of the coefficients obtained with Ridge Regression revealed that the most relevant wavelengths for predicting C content are primarily concentrated in the $1300$ to $1400$ nm range. Among these, $1346$ nm, $1349$ nm, $1344$ nm, and $1353$ nm stand out, as they exhibited the largest absolute coefficient values, both positive and negative. This indicates that different bands in this region contribute distinctively to the model, with some increasing and others reducing the prediction. These findings are coherent with the NIR spectroscopy literature, which associates this region with absorptions linked to organic functional groups, such as C–H and O–H, which are directly related to soil C. Beyond this key range, other wavelengths, such as $947$ nm, $1173$ nm, $1547$ nm, and $1564$ nm, also showed relevance, albeit with a smaller magnitude.

Conversely, the least important wavelengths for the model were found mainly around $1600–1700$ nm (e.g., $1673$, $1676$, and $1678$ nm) and near $950–1000$ nm ($965$, $975$, and $1009$ nm). These regions showed very low coefficients, suggesting little to no contribution to C prediction. It is likely that such bands are more associated with experimental noise or absorptions from soil constituents that are not directly related to C, such as water or minerals.

The analysis of the Ridge model coefficients for predicting nitrogen shows that the most relevant wavelengths are again concentrated in the region between $1300$ and $1350$ nm (especially $1344$, $1346$, $1340$, and $1349$ nm), with a strong positive contribution. This spectral range is already well-known for reflecting absorptions of functional groups related to N–H, O–H, and C–H bonds, which justifies its importance for soil nitrogen modeling. Additionally, bands near $1170$ nm ($1173$, $1169$, $1167$ nm) also appeared as important, reinforcing the model's reliance on specific regions of the mid-NIR spectrum.

Among the least relevant wavelengths for nitrogen, regions around $1600–1700$ nm ($1640$, $1670$, $1673$, $1676$, $1678$ nm), as well as wavelengths near $970–1000$ nm ($971$, $981$, $1037$ nm), stand out. These values presented very low coefficients, suggesting a limited ability to discriminate nitrogen content.In summary, this study reinforces the importance of the mid-NIR region ($1300–1400$ nm) in estimating C and N content and identifies wavelengths that could be discarded in future analyzes. This selection process could aid in reducing the dimensionality of the spectrum, which, in turn, paves the way for the implementation of potential variable selection algorithms.

\begin{table}[!h]
\centering
\caption{Main wavelengths selected by the Ridge model for Carbon and Nitrogen prediction in Oxisol.}
\label{tab:ridge_features_cn}
\renewcommand{\arraystretch}{1.3}
\begin{tabular}{cccc}
\hline
\multicolumn{2}{c}{\textbf{Carbon}} & \multicolumn{2}{c}{\textbf{Nitrogen}} \\ 
\cline{1-4}
\textbf{20 most relevant} & \textbf{20 least relevant} & \textbf{20 most relevant} & \textbf{20 least relevant} \\ 
\hline
\cellcolor{green!10}1346 & \cellcolor{red!10}1515 & \cellcolor{green!10}1344 & \cellcolor{red!10}1527 \\ \hline
\cellcolor{green!10}1349 & \cellcolor{red!10}1673 & \cellcolor{green!10}1346 & \cellcolor{red!10}1640 \\ \hline
\cellcolor{green!10}1344 & \cellcolor{red!10}1009 & \cellcolor{green!10}1340 & \cellcolor{red!10}1238 \\ \hline
\cellcolor{green!10}1353 & \cellcolor{red!10}965 & \cellcolor{green!10}1173 & \cellcolor{red!10}1203 \\ \hline
\cellcolor{green!10}1610 & \cellcolor{red!10}975 & \cellcolor{green!10}1169 & \cellcolor{red!10}1678 \\ \hline
\cellcolor{green!10}1169 & \cellcolor{red!10}1681 & \cellcolor{green!10}1167 & \cellcolor{red!10}1670 \\ \hline
\cellcolor{green!10}1167 & \cellcolor{red!10}1203 & \cellcolor{green!10}1349 & \cellcolor{red!10}1414 \\ \hline
\cellcolor{green!10}1356 & \cellcolor{red!10}1078 & \cellcolor{green!10}1163 & \cellcolor{red!10}1530 \\ \hline
\cellcolor{green!10}1435 & \cellcolor{red!10}1629 & \cellcolor{green!10}1337 & \cellcolor{red!10}1199 \\ \hline
\cellcolor{green!10}1432 & \cellcolor{red!10}1447 & \cellcolor{green!10}1312 & \cellcolor{red!10}1277 \\ \hline
\cellcolor{green!10}1340 & \cellcolor{red!10}1290 & \cellcolor{green!10}1547 & \cellcolor{red!10}1618 \\ \hline
\cellcolor{green!10}1429 & \cellcolor{red!10}1331 & \cellcolor{green!10}1037 & \cellcolor{red!10}1328 \\ \hline
\cellcolor{green!10}1173 & \cellcolor{red!10}1244 & \cellcolor{green!10}1510 & \cellcolor{red!10}1376 \\ \hline
\cellcolor{green!10}1163 & \cellcolor{red!10}936 & \cellcolor{green!10}1123 & \cellcolor{red!10}981 \\ \hline
\cellcolor{green!10}1312 & \cellcolor{red!10}1450 & \cellcolor{green!10}1353 & \cellcolor{red!10}1676 \\ \hline
\cellcolor{green!10}1358 & \cellcolor{red!10}1592 & \cellcolor{green!10}1176 & \cellcolor{red!10}1626 \\ \hline
\cellcolor{green!10}947 & \cellcolor{red!10}1676 & \cellcolor{green!10}1429 & \cellcolor{red!10}1503 \\ \hline
\cellcolor{green!10}1547 & \cellcolor{red!10}1678 & \cellcolor{green!10}1550 & \cellcolor{red!10}971 \\ \hline
\cellcolor{green!10}1564 & \cellcolor{red!10}1416 & \cellcolor{green!10}1553 & \cellcolor{red!10}1576 \\ \hline
\cellcolor{green!10}1283 & \cellcolor{red!10}1589 & \cellcolor{green!10}1159 & \cellcolor{red!10}1673 \\ \hline
\end{tabular}
\end{table}

Table \ref{tab:ridge_features_cn} summarizes the band regions, in nanometers (nm), corresponding to the C and N contents. Meanwhile, Figure \ref{fig:ridge_map} displays only the spectrum of the Oxisol samples in relation to the estimated N values, given that the graph for C exhibits high similarity. The figure is shown in two colors: green for the $70\%$ of the most relevant regions and red for the $30\%$ of the least relevant regions for the Ridge model. All spectral regions still maintain some contribution to the final model prediction. The regions shown in red are, therefore, merely the least impactful.

\begin{figure}[!h]
\includegraphics[width=1.0\textwidth]{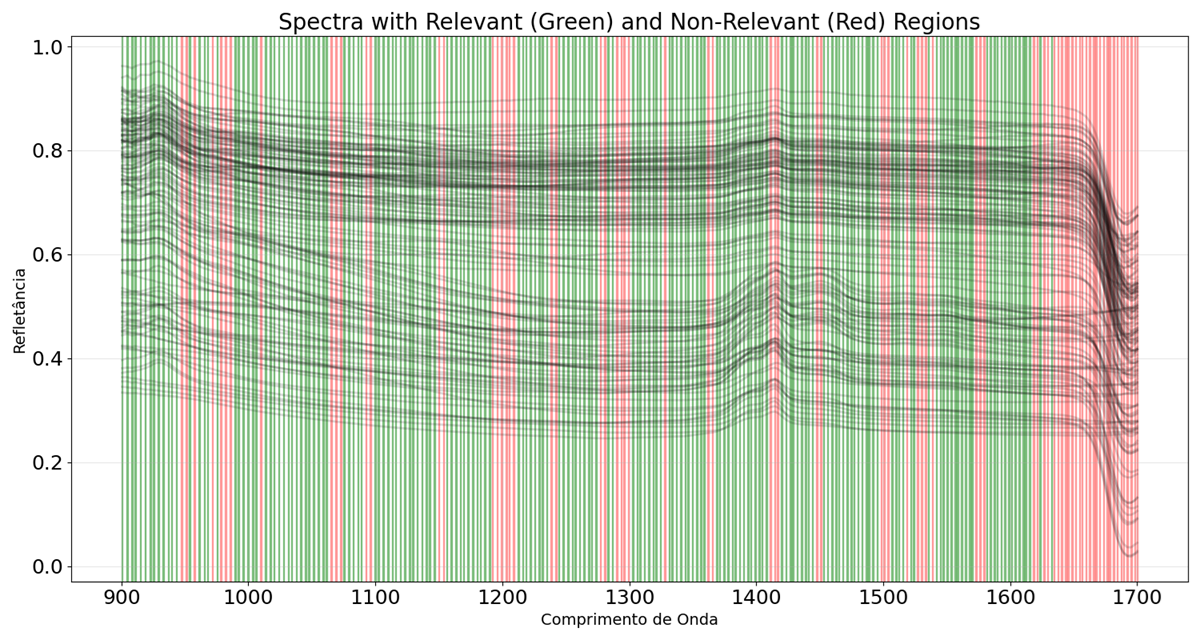}
\caption{Spectra with Relevant (Green) and Non-Relevant (Red) Regions}
\label{fig:ridge_map}
\end{figure}

\section{Conclusion}\label{sec:conclusion}

This work successfully compared various strategies for NIR reflectance spectral data preprocessing and multiple ML algorithms to predict C and N contents in two important Brazilian soil types: Oxisol and Inceptisol. The results confirm that NIR spectroscopy is a viable, rapid, and non-destructive analytical tool, emphasizing that both preprocessing techniques and the choice of the appropriate pipelines significantly influence the optimization of the prediction results.

The most effective preprocessing strategy identified was the combination of the SG filter with outlier removal, which proved superior for both soil types and both components (C and N). A key finding was the clear influence of soil type on model performance; the models performed better for Oxisol. This is likely due to Oxisol's intrinsic pedological characteristics and potentially the wider range and greater variability of its C and N contents, which provided a more representative dataset for model training.

Regarding data splitting, the study highlighted that the statistical representativeness of the test set is critical. The most suitable splits were the Holdout 70/30 (Oxisol) and 75/25 (Inceptisol), both utilizing Kennard-Stone sampling. The research also employed and compared robust error estimation strategies, including Cross-Validation with 10 Folds and Leave-one-Out.

The optimal predictive models varied by soil type. For Oxisol, pipeline 2 (SVR with Data Standardization) demonstrated the best generalization capacity for both components, achieving a strong $R^2$ of $0.91$ and an $RPD$ of $3.39$ for C, and an $R^2$ of $0.89$ and an $RPD$ of $3.01$ for N. In contrast, Inceptisol samples required different models: pipeline 6 (Stacking Ensemble of PLS with Ridge) was best for C ($R^2=0.79$, $RPD=2.16$), while pipeline 4 (PLS) was best for N ($R^2=0.77$, $RPD=2.11$). Generally, Stacking Ensemble models tended to produce more robust and consistent results across calibration and test sets, with the addition of SVR proving beneficial, especially for N in Oxisol. RF, conversely, consistently showed overfitting across all scenarios, likely due to the limited sample size available for this specific model in the study, while PLS maintained an intermediate performance level.

\subsection{Main Contributions}\label{subsec:main_contributions}

 This study impacts the areas of chemometrics and precision agriculture by developing and optimizing soil-specific models, with SVR and PLS established as the most robust options for Oxisol and Inceptisol, respectively. The most important spectral regions analysis via the ridge model may help low-cost systems by targeting those regions to acquire information. The study reaveals that the mid-NIR region (1300 to 1400 nm) is the most relevant spectral range for C and N prediction in both Oxisol and Inceptisol, confirming the significance of absorptions related to organic functional groups. On the other hand, the 1600 to 1700 nm region was the least influential. In conclusion, the findings may result in more informed decision-making by producers and consultants regarding organic matter, fertility, and nutrient availability.

\subsection{Future Work}\label{subsec:main_contributions}

The research's next steps include the implementation of variable selection algorithms within the spectral regions identified by Ridge Regression to simplify the models. Future work also involves developing multivariate systems to predict diverse soil properties simultaneously, aiming to abstract these concepts to other application areas and improve user interaction.

\section*{CrediT authorship contribution statement}

\textbf{Vinicius H. Kieling}: writting - original draft; investigation; methodology; conceptualization; software. \textbf{Felipe A. B. Rossi}: data curation; investigation; writing. \textbf{Guilherme Baggio}: writing; investigation; software. \textbf{Marco A. de C. Barbosa}: validation, supervision, review. \textbf{Dalcimar Casanova}:validation, supervision, review.  \textbf{Larissa M. dos Santos-Tonial}: review; data curation; validation, supervision. \textbf{Jefferson T. Oliva}: investigation; validation, supervision, review, editing.

\section*{Declaration of competing interest}

We declare that there is no known conflict of interest related to this work.

\section*{Ethics Statement}
Not applicable: This manuscript does not include human or animal research

\section*{Declaration of generative AI use}

The authors used Gemini and Grammarly for text review to correct spelling errors. All methodological decisions, interpretation results, experimental validation, and final drafting of the work remaining the responsibility of the authors.

\section*{Data availability}

Data will be made available on request.

\section*{Acknowledgements}

The authors would like to thank the Fundação Araucária (Edital 09/2021, grant number PBA2022011000259) for the financial support, and the Central de Análises (UTFPR-PB) for the support analysis.

\bibliography{nir}

\end{document}